\documentclass[twoside]{article}

\usepackage[utf8]{inputenc} %
\usepackage[T1]{fontenc} %
\usepackage{hyperref} %
\usepackage{url} %
\usepackage{booktabs} %
\usepackage{amsfonts} %
\usepackage{nicefrac} %
\usepackage{microtype} %

\usepackage{enumerate}
\usepackage{natbib}
\usepackage{hyperref}
\usepackage{multirow}
\usepackage{multicol}
\usepackage{makecell}
\usepackage{amsmath, amsthm, amssymb,graphicx}
\usepackage[ruled,longend]{algorithm2e}
\usepackage{xcolor}
\usepackage{appendix}
\usepackage{caption}
\usepackage{subcaption}

\usepackage{balance}

\usepackage{marginnote}

\DeclareMathOperator*{\argmin}{arg\,min}

\usepackage{stackengine}

\makeatletter
\newcommand{\distas}[1]{\mathbin{\overset{#1}{\kern\z@\sim}}}%

\newtheorem{theorem}{Theorem}

\newtheorem{cor}[theorem]{Corollary}

\newcommand{\bas}[1]{\begin{align*}#1\end{align*}}
\newcommand{\ba}[1]{\begin{align}#1\end{align}}

\newcommand{\bphi}{\boldsymbol{\phi}}

\newcommand{\beqs}{\vspace{0mm}\begin{eqnarray}}
\newcommand{\eeqs}{\vspace{0mm}\end{eqnarray}}
\newcommand{\barr}{\begin{array}}
\newcommand{\earr}{\end{array}}

\newcommand{\av}[0]{{\boldsymbol{a}}}

\newcommand{\dv}{\boldsymbol{d}}

\newcommand{\gv}[0]{{\boldsymbol{g}} }

\newcommand{\sv}[0]{{\boldsymbol{s}}}

\newcommand{\cdotv}{\boldsymbol{\cdot}}

\newcommand{\Pimat}[0]{{\boldsymbol{\Pi}} }

\newcommand{\thetav}{\boldsymbol{\theta}}

\newcommand{\phiv}{\boldsymbol{\phi}}

\newcommand{\varpiv}[0]{{\boldsymbol{\varpi}} }

\newcommand{\omegav}[0]{{\boldsymbol{\omega}}}

\newcommand{\E}{\mathbb{E}}

\newcommand{\given}{\,|\,}

\newcommand{\yy}[1]{{\color{blue}{[yuguang: #1]}}}
 
\newcommand{\rr}[1]{\textcolor{red}{mz:#1}} 

 \usepackage[accepted]{aistats2020}

\begin{document}

\twocolumn[

\aistatstitle{%
Discrete Action On-Policy Learning with Action-Value Critic
}

\aistatsauthor{Yuguang Yue \And Yunhao Tang \And Mingzhang Yin \And Mingyuan Zhou}

\aistatsaddress{ UT-Austin \And Columbia University \And UT-Austin \And UT-Austin} ]

\begin{abstract}

 Reinforcement learning (RL) in discrete action space is ubiquitous in real-world applications, but its complexity grows exponentially with the action-space dimension, making it challenging to apply existing on-policy gradient based deep RL algorithms efficiently. To effectively operate in multidimensional discrete action spaces, we construct a critic to estimate action-value functions, apply it on correlated actions, and combine these critic estimated action values to control the variance of gradient estimation. We follow rigorous statistical analysis to design how to generate and combine these correlated actions, and how to sparsify the gradients by shutting down the contributions from certain dimensions. These efforts result in a new discrete action on-policy RL algorithm that empirically outperforms related on-policy algorithms relying on variance control techniques. We demonstrate these properties on OpenAI Gym benchmark tasks, and illustrate how discretizing the action space could benefit the exploration phase and hence facilitate convergence to a better local optimal solution thanks to the flexibility of discrete policy. 
 
%
%
%
%
%
 
 %

\end{abstract}

\section{\uppercase {Introduction}}\label{intro}

There has been significant recent interest in using model-free reinforcement learning (RL) %
to address complex real-world sequential decision making tasks \citep{silver2018general,macalpine2017ut,OpenAI_dota}. With the help of deep neural networks, model-free deep RL algorithms have been successfully implemented in a variety of tasks, including game playing \citep{silver2016mastering, mnih2013playing} and robotic controls \citep{levine2016end}. Among those model-free RL algorithms, policy gradient (PG) algorithms are a class of methods that parameterize the policy function and apply gradient-based methods to make updates. It has been shown to succeed in solving a range of challenging RL tasks \citep{mnih2016asynchronous,schulman2015trust,lillicrap2015continuous,schulman2017proximal,wang2016sample,haarnoja2018soft,liu2017stein}. Despite directly targeting at %
maximizing the expected rewards, PG suffers from problems including having low sample efficiency \citep{haarnoja2018soft} for on-policy PG algorithms 
and undesirable sensitivity to hyper-parameters for off-policy algorithms \citep{lillicrap2015continuous}.

On-policy RL algorithms use on-policy samples to estimate the gradients for policy parameters, as routinely approximated by Monte Carlo (MC) estimation that often comes %
with large variance. A number of techniques have sought to alleviate this problem for continuous action spaces \citep{gu2016q,grathwohl2017backpropagation,liu2017action,wu2018variance}, while relatively fewer have been proposed for discrete action spaces \citep{grathwohl2017backpropagation,ARSM}. 
 In practice, RL with discrete action space is ubiquitous in fields including recommendation system \citep{dulac2015deep}, bidding system \citep{hu2018reinforcement}, gaming \citep{mnih2013playing}, to name a few. 
It plays an important role in the early stage of RL development \citep{sutton2018reinforcement}, and many value-based algorithms \citep{watkins1992q,mnih2013playing,van2016deep} can handle such setup when the action space is not large. However, when the action space is multidimensional, the number of unique actions grows exponentially with the dimension,
leading to an intractable combinatorial optimization problem at every single step that prevents %
the application of most value-based RL methods.

Under the setting of high-dimensional discrete action space, policy-gradient based algorithms can still be applied if we assume the joint distribution over discrete actions to be
factorized across dimensions, so that the joint policy is still
tractable 
\citep{jaskowski2018reinforcement,andrychowicz2018learning}.
Then the challenge boils down to obtaining a gradient estimator that can well control its variance.
Though many variance reduction techniques have been proposed for discrete variables \citep{jang2016categorical,tucker2017rebar,ARM,raiko2014techniques}, they either provide biased gradients or are not applicable to multidimensional RL settings. 

In this paper, we propose Critic-ARSM (CARSM) policy gradient, which improves the recently proposed augment-REINFORCE-swap-merge (ARSM) gradient estimator of \citet{ARSM} and integrates it with %
action-value function evaluation, to accomplish three-fold effects: \textbf{1)} CARSM sparsifies the %
ARSM %
gradient and introduces an action-value Critic to work with %
multidimensional discrete actions spaces; %
{\textbf{2)}
 By estimating the rewards of a set of correlated discrete actions via the proposed action-value Critic, and combining these rewards for variance reduction, %
CARSM 
achieves better sample efficiency compared with other variance-control methods such as A2C \citep{mnih2016asynchronous} and RELAX \citep{grathwohl2017backpropagation}}; 
 \textbf{3)} CARSM can be easily applied to other RL algorithms using REINFORCE or its variate as the gradient estimator. Although we mainly focus on on-policy algorithms, our algorithm can also be potentially applied to off-policy algorithms with the same principle; we leave this extension for future study. 

The paper proceeds as follows. In Section 2, we briefly review existing on-policy learning frameworks %
and variance reduction techniques for discrete action space. In Section 3, we introduce CARSM from both theoretical and practical perspectives. In Section 4, we first demonstrate the potential benefits of discretizing a continuous control task compared with using a diagonal Gaussian policy, then show the high sample efficiency of CARSM from an extensive range of experiments and illustrate 
that CARSM can be plugged into state-of-arts on-policy RL learning frameworks such as Trust Region Policy Optimization (TRPO) \citep{schulman2015trust}. Python (TensorFlow) code is available  at  {\url{https://github.com/yuguangyue/CARSM}.}

\section{\uppercase{Preliminaries}}\label{prelim}
RL is often formulated as learning %
under a Markov decision process (MDP). Its action space $\mathcal{A}$ is dichotomized into either discrete ($e.g.$, $\mathcal{A} = \{1,\ldots,100\}$) or continuous ($e.g.$, $\mathcal{A} = [-1,1]$). In an MDP, at discrete time $t\geq 0$, an agent in state $\sv_t \in \mathcal{S}$ takes action $a_t \in \mathcal{A}$, receives instant reward $r(\sv_t,a_t)\in \mathbb{R}$, and transits to next state $\sv_{t+1} \sim \mathcal P(\cdotv\given \sv_t,a_t)$. Let $\pi:\mathcal{S} \mapsto\mathcal{P}(\mathcal{A})$ be a mapping from the state to a distribution over actions. We define the expected cumulative rewards under $\pi$ as
\begin{equation}\label{obj}
J(\pi) \textstyle = \mathbb{E}_{\pi}\left[\sum_{t = 0}^\infty\gamma^t r(\sv_t,a_t)\right], %
\end{equation}
where $\gamma \in (0,1]$ is a discount factor. The objective of RL is to find the (sub-)optimal policy $\pi^\ast = \arg\max_\pi J(\pi)$. 
In practice, it is infeasible to search through all policies and hence one typically resorts to parameterizing the policy $\pi_{\thetav}$ with $\thetav$. %

\subsection{On-Policy Optimization}
We introduce on-policy optimization methods from a constrained optimization point of view to unify the algorithms we will discuss in this article. In practice, we want to solve the following constrained optimization problem as illustrated in \citet{schulman2015trust}:
\begin{equation*}
 \begin{aligned}
 &\textstyle\max_{\thetav}\ \mathbb{E}_{\pi_{\thetav_{\text{old}}}} \Big[\frac{\pi_{\thetav}(a_t|\sv_t)}{\pi_{\thetav_{\text{old}}} (a_t|\sv_t)} Q^{\pi_{\thetav_{\text{old}}}}(\sv_t,a_t)\Big]\\
 &~~~~~~~~\text{subject to}\ D(\thetav_{\text{old}},\thetav) \leq\epsilon,
 \end{aligned}
\end{equation*}
where $Q^{\pi_{\thetav}}(\sv_t,a_t) = \mathbb{E}_{\pi_{\thetav}}[\sum_{t'=t} \gamma^{t'-t} r(s_t',a_t')]$ is the action-value function and $D(\cdot,\cdot)$ is some metric that measures the closeness between $\thetav_{\text{old}}$ and $\thetav$. 

\textbf{A2C Algorithm: }
One choice of $D(\cdot,\cdot)$ is the $L_2$ norm, which will lead us to first-order gradient ascent. By applying first-order Taylor expansion on $\pi_{\thetav}$ around $\thetav_{\text{old}}$, the problem can be re-written as maximizing $\mathbb{E}_{\pi_{\thetav_{\text{old}}}} [Q^{\pi_{\thetav_{\text{old}}}}(\sv_t,a_t)] + \nabla_{\thetav} J(\pi_{\thetav_{\text{old}}})^T (\thetav-\thetav_{\text{old}})$ subject to $||\thetav-\thetav_{\text{old}}||_2\leq\epsilon$, which will result in a gradient ascent update scheme; note $\nabla_{\thetav} J(\pi_{\thetav_{\text{old}}}):=\nabla_{\thetav} J(\pi_{\thetav})|_{\thetav = \thetav_{\text{old}}}$.
Based on REINFORCE \citep{williams1992}, the gradient of the original objective function (\ref{obj}) can be written as
\begin{equation}\label{REINFORCE}
 \small \textstyle \nabla_{\thetav}J(\pi_{\thetav}) = \mathbb{E}_{\pi_{\thetav}} \left[\sum_{t=0}^\infty Q^{\pi_{\thetav}}(\sv_t,a_t) \nabla_{\thetav}\log\pi_{\thetav}(a_t|\sv_t)\right]. %
\end{equation}
 However, a naive Monte Carlo estimation of (\ref{REINFORCE}) has large variance that needs to be controlled. 
 A2C algorithm \citep{mnih2016asynchronous} adds value function $V^{\pi_{\thetav}}(\sv):=\E_{a_t\sim\pi_{\theta}}[Q^{\pi_{\thetav}}(\sv_t,a_t)]$ as a %
 baseline and obtains a low-variance estimator of $\nabla_{\thetav}J(\pi_{\thetav})$ as
 \begin{equation}\label{ga2c}
\textstyle \gv_{\text{A2C}} =\mathbb{E}_{\pi_{\thetav}} \left[\sum_{t=0}^\infty A^{\pi_{\thetav}}(\sv_t,a_t) \nabla_{\thetav}\log\pi_{\thetav}(a_t|\sv_t)\right],
 \end{equation}
 where $A^{\pi_{\thetav}}(\sv_t,a_t) = Q^{\pi_{\thetav}}(\sv_t,a_t) - V^{\pi_{\thetav}}(\sv_t)$ is called the Advantage function. 

\textbf{Trust Region Policy Optimization: }
The other choice of metric $D(\cdot,\cdot)$ could be KL-divergence, and the update from this framework is introduced as TRPO \citep{schulman2015trust}. In practice, this constrained optimization problem is reformulated as follows:
\begin{equation}\label{trpo_max}
\begin{aligned}
 &\max_{\thetav}\ \nabla_{\thetav} J(\pi_{\thetav_{\text{old}}})^T (\thetav - \thetav_{\text{old}}) \\
 &\text{subject to} \textstyle \ \frac{1}{2}(\thetav_\text{old} - \thetav)^T H(\thetav_\text{old} - \thetav)\leq \delta,
\end{aligned}\nonumber
\end{equation}
where $H$ is the second-order derivative $\nabla_{\thetav}^2D_{\text{KL}}(\thetav_{\text{old}}||\thetav)|_{\thetav = \thetav_{\text{old}}}$. An analytic update step for this optimization problem can be expressed as 
\begin{equation}\label{trpo}
\textstyle \thetav = \thetav_{\text{old}} + \sqrt{\frac{2\delta}{\dv^TH^{-1}\dv}}\dv,
\end{equation}
where $\dv = H^{-1}\nabla_{\thetav}J(\pi_{\thetav_{\text{old}}})$, and in practice the default choice of $\nabla_{\thetav}J(\pi_{\thetav_{\text{old}}})$ is $\gv_{\text{A2C}}$ as defined at (\ref{ga2c}).

\subsection{Variance Control Techniques}
Besides the technique of using state-dependent baseline to reduce variance as %
in (\ref{ga2c}), two recent works propose alternative methods for variance reduction in %
discrete action space settings. For the sake of space, we defer to \citet{grathwohl2017backpropagation} for the detail about the RELAX algorithm and briefly introduce ARSM here. 

\textbf{ARSM Policy Gradient: }
The ARSM gradient estimator can be used to backpropagate unbiased and low-variance gradients through a sequence of unidimensional categorical variables \citep{ARSM}. It comes up with a reparametrization formula for discrete random variable, and combines it with a parameter-free self-adjusted baseline to achieve variance reduction. %

Instead of manipulating on policy parameters $\thetav$ directly, ARSM turns to reduce variance on the gradient with respect to the logits $\phiv$, before backpropagating it to $\thetav$ using the chain rule. 
Let us assume $$\pi_{\thetav}(a_t\given \sv_t) = \mbox{Categorical}(a_t\given \sigma(\phiv_t)),~\phiv_t:=\mathcal{T}_{\thetav}(\sv_t),$$
where $\sigma(\cdotv)$ denotes the softmax function and $\mathcal{T}_{\thetav}(\cdotv)\in\mathbb{R}^C$ denotes a neural network, which is parameterized by $\thetav$ and has an output dimension of $C$.

Denote $\varpiv^{_{{c} \leftrightharpoons j}}$ as the vector obtained by swapping the $c$th and $j$th elements of vector $\varpiv$, which means $\varpi_j^{_{{c} \leftrightharpoons j}}=\varpi_c$, $\varpi_c^{_{{c} \leftrightharpoons j}}=\varpi_j$, and $\varpi_i^{_{{c} \leftrightharpoons j}}=\varpi_i$ if $i\notin \{c,j\}$.
Following the derivation from \citet{ARSM}, the gradient with respect to $\phi_{tc}$ can be expressed as
\bas{
&\small \nabla_{\phi_{tc}}J(\phiv_{0:\infty})%
 = \E_{\mathcal{P}(\sv_{t}\given \sv_0,\pi_{\thetav}) \mathcal{P}(\sv_{0}) }\left\{\gamma^{t} \E_{\varpiv_t \sim \text{Dir}(\mathbf{1}_C) }
\left[g_{tc}\right]\right\},\notag\\
&\resizebox{0.94\hsize}{!}{$\displaystyle g_{tc}:=\sum_{j=1}^C \left[Q(\sv_t,a_t^{c\leftrightharpoons j})-\frac{1}{C}\sum_{m=1}^C Q(\sv_t,a_t^{m\leftrightharpoons j})\right]\left(\frac{1}{C}-\varpi_{tj}\right)$},%
}
where $\mathcal{P}(\sv_{t}\given \sv_0,\pi_{\thetav}) $ is the marginal form of 
$ \prod_{t'=0}^{t-1}\mathcal{P}(\sv_{t'+1}\given \sv_{t'},a_{t'}) \text{Categorical}(a_{t'};\sigma(\phiv_{t'}))
$, $\varpiv_t\sim\text{Dir}(\mathbf{1}_C)$, and $a_t^{_{{c} \leftrightharpoons j}}: = \argmin_{i\in\{1,\ldots,C\}} \varpi_{ti}^{_{{c} \leftrightharpoons j}} e^{-\phi_{ti}}$. In addition, $a_t^{_{{c} \leftrightharpoons j}}$ is called a \textit{pseudo action} to differentiate it from the true action $a_t=: \argmin_{i\in\{1,\ldots,C\}} \varpi_{ti}e^{-\phi_{ti}}$.

Applying the chain rule leads to ARSM policy gradient:
\begin{equation*}
\begin{aligned}
&{\gv}_{\text{ARSM}} = \textstyle\sum_{t=0}^\infty\sum_{c=1}^C \frac{\partial J(\phiv_{0:\infty})}{\partial\phi_{tc}}\frac{\partial\phi_{tc}}{\partial\thetav}\\
&\textstyle= \mathbb{E}_{\sv_t\sim \rho_{\pi,\gamma}(\sv) }\left\{ \E_{\varpiv_t\sim \text{Dir}(\mathbf{1}_C) }
 \left[ \nabla_{\thetav} \sum_{c=1}^C g_{tc} \phi_{tc}\right] 
\right\},
\end{aligned}
\end{equation*}
where $\rho_{\pi,\gamma}(\sv):= \!\sum_{t=0}^\infty \gamma^t \mathcal{P}(\sv_t=\sv \given \sv_0,\pi_{\thetav})$ is the unnormalized discounted state visitation frequency. 

In \cite{ARSM}, $Q(\sv_t,a_t^{c\leftrightharpoons j})$ are estimated by MC integration, which requires multiple MC rollouts at each timestep if there are \textit{pseudo actions} that differ from the true action. This estimation largely limits the implementation of ARSM policy gradient %
to small action space due to the high %
computation cost. The maximal number of unique \textit{pseudo actions} grows quadratically with the number of actions along each dimension and a long episodic task will result in more MC rollouts too. To differentiate it from the new algorithm, we refer to it as ARSM-MC. 

\section{\uppercase {CARSM Policy Gradient}}\label{arsmc}
In this section, we introduce Critic-ARSM (CARSM) policy gradient for multidimensional discrete action space. CARSM improves ARSM-MC in the following two aspects: 1. ARSM-MC only works for unidimensional RL settings while CARSM generalizes it to multidimensional ones with sparsified gradients. %
2. CARSM can be applied to more complicated tasks as it employs an action-value function critic to remove the need of running multiple MC rollouts for a single estimation, which largely improves the sample efficiency. 

For an RL task with $K$-dimensional $C$-way discrete action space, we assume different dimensions $a_{tk}\in\{1,\ldots,C\}$ of the multidimensional discrete action $\av_t=(a_{t1},\ldots,a_{tK})$ are independent given logits $\phiv_t$ at time $t$, that is
$ %
a_{t1} \bot a_{t2}\cdots \bot a_{tK}\given \phiv_t\notag
$. For the logit vector $\phiv_t\in\mathbb{R}^{KC}$, which can be decomposed as $\phiv_t =(\phiv_{t1}',\ldots,\phiv_{tK}')',~\phiv_{tk}=(\phi_{tk1},\ldots,\phi_{tkC})'$, we assume 
\bas{P(\av_t\given \phiv_t)=\textstyle\prod_{k=1}^K \mbox{Categorical}(a_{tk};\sigma(\phiv_{tk})).
} %

\begin{theorem}
[Sparse ARSM for multidimensional discrete action space]\label{thm1}

The element-wise gradient of $J(\phiv_{0:\infty})$ with respect to $\phi_{tkc}$ can be expressed as 
\bas{
\resizebox{0.99\hsize}{!}{$
\textstyle \nabla_{\phi_{tk{c}}}J(\bphi_{0:\infty}) = %
 \E_{\mathcal{P}(\sv_{t}\given \sv_0,\pi_{\thetav}) \mathcal{P}(\sv_{0}) }
 \left\{\gamma^t
 \E_{ \Pimat_t\sim
 \prod_{{k}=1}^{{K}}{\emph{\text{Dir}}}(\varpiv_{tk};\mathbf{1}_{C})}
 [g_{tkc}]
 \right\},
$}
}
where $\varpiv_{tk}=(\varpi_{tk1},\ldots,\varpi_{tkC})'\sim\emph{\mbox{Dir}}(\mathbf{1}_C)$ is the Dirichlet random %
vector for dimension $k$, state $t$ and
\bas{
&g_{tkc} :=
 \left\{
 \begin{array}{ll}
 0,\qquad\qquad \text{if } a_{tk}^{_{{c} \leftrightharpoons j}} = a_{tk}\text{ for all (c,j)}\\
 \sum_{j=1}^C \left[\Delta_{c,j}(\sv_t,\av_t)\right] \left(\frac{1}{C}-\varpi_{tkj}\right),\ \text{otherwise}\notag\\
 \end{array}
 \right.\\
&\textstyle \Delta_{c,j}(\sv_t,\av_t):=Q(\sv_t,\av_t^{_{{c} \leftrightharpoons j}} ) - \frac{1}{C}\sum_{m=1}^{C} Q(\sv_t,\av_t^{_{{m} \leftrightharpoons j}} ), \\
& {\av}_{t}^{_{{c} \leftrightharpoons j}}: =( a_{t1}^{_{{c} \leftrightharpoons j}},\ldots, a_{tK}^{_{{c} \leftrightharpoons j}})', \\
&a_{tk}^{_{{c} \leftrightharpoons j}}:\textstyle=\argmin_{i\in\{1,\ldots,C\}} \varpi_{tki}^{_{c \leftrightharpoons j}}e^{-\phi_{tki}}. 
}
\end{theorem}

 We defer the proof to Appendix \ref{proof_thm1}. One difference from the original ARSM \citep{ARSM} is the values of %
 $g_{tkc}$, where we obtain a sparse estimation that shutdowns the $k$th dimension if $a_{tk}^{_{{c} \leftrightharpoons j}} = a_{tk}$ for all $(c,j)$ and hence there is no more need to %
 calculate $\nabla_{\thetav}\phi_{tkc}$ for all $c$ belonging to dimension $k$ at time $t$. One immediate benefit from this sparse gradient estimation is to reduce the noise from that specific dimension because the $Q$ function is always estimated with either MC estimation or Temporal Difference (TD) \citep{sutton2018reinforcement}, which will introduce variance and bias, respectively.
 %

In ARSM-MC, the action-value function is estimated by MC rollouts. Though it returns unbiased estimation, it inevitably decreases the sample efficiency and prevents it from applying to more sophisticated tasks. Therefore, CARSM proposes using an action-value function critic $\hat{Q}_{\omegav}$ parameterized by $\omegav$ to estimate the $Q$ function. 
Replacing $Q$ with $\hat{Q}_{\omegav}$ in Theorem~\ref{thm1}, we obtain $\hat{g}_{{tkc}}\approx g_{tkc}$ as the empirical estimation of $\nabla_{\phi_{tkc}} J(\pi)$, 
and hence %
the CARSM estimation for $\nabla_{\thetav} J(\pi)=\sum_{t=0}^\infty\sum_{k=1}^K\sum_{c=1}^C \frac{\partial J(\phiv_{0:\infty})}{\phi_{tkc}}\frac{\phi_{tkc}}{\partial\thetav}$ becomes %
\bas{\textstyle
\hat{\gv}_{\text{CARSM}} = \nabla_{\thetav} \sum_{t}\sum_{k=1}^K\sum_{c=1}^C \hat{g}_{tkc} \phi_{tkc}.
}
Note %
the number of unique %
values in $\{{\av}_{t}^{_{{c} \leftrightharpoons j}}\}_{c,j}$ that differ from the true action $\av_t$ is always between $0$ and $C(C-1)/2-1$, regardless of how large $K$ is. The dimension shutdown property further sparsifies the gradients, %
removing %
the noise of the dimensions that have no \textit{pseudo actions}.

\textbf{Design of Critic: } 
A practical challenge of CARSM is that it is notoriously hard to estimate action-value functions for on-policy algorithm because the number of samples are limited and the complexity of the action-value function quickly increases with dimension $K$. A natural way to overcome the limitation of samples is the reuse of historical data, which has been successfully implemented in previous studies \citep{gu2016q,lillicrap2015continuous}. The idea is to use the transitions $\{\sv_{\ell},r_{\ell},\av_{\ell},\sv_{\ell}'\}$'s from the replay buffer to construct target values for the action-value estimator under the current policy. More specifically, we can use one-step TD to rewrite the target value of critic $\hat{Q}_{\omegav}$ network with these off-policy samples as 
\begin{equation}\label{Q_target}
y^{\text{off}}_{\ell} = r(\sv_{\ell},\av_{\ell}) + \gamma\mathbb{E}_{\tilde{\av}\sim\pi(\cdotv \given \sv_{\ell}')} \hat{Q}_{\omegav}(\sv_{\ell}', \tilde{\av}),
\end{equation}
where the expectation part can be evaluated with either an exact computation when the action space size $C^K$ is not large, or with MC integration by drawing random samples from $\tilde{\av}\sim\pi_{\thetav}(\cdotv \given \sv')$ and averaging $\hat{Q}_{\omegav}(\sv', \tilde{\av})$ over these random samples. 
This target value only uses one-step estimation, and can be extended to $n$-step TD %
by adding additional importance sampling weights. 

Since we have on-policy samples, it is natural to also include them to construct unbiased targets for $\hat{Q}_{\omegav}(\sv_t,a_t)$:
\begin{equation*}
 y^{\text{on}}_t = \textstyle \sum_{t'=t}^\infty \gamma^{t'-t} r(\sv_{t'},\av_{t'}).
\end{equation*}
Then we optimize parameters $\omegav$ by minimizing the Bellman error between the targets and critic as
\begin{equation*}
\textstyle
 \sum_{\ell=0}^L [y_{\ell}^{\text{off}} - \hat{Q}_{\omegav}(\sv_{\ell}, \av_{\ell})]^2+\sum_{t=0}^T [y_{t}^{\text{on}} - \hat{Q}_{\omegav}(\sv_{t}, \av_{t})]^2,
\end{equation*}
where $L$ is the number of off-policy samples and $T$ is the number of on-policy samples. 
In practice, the performance varies with the ratio between $L$ and $T$, which reflects the trade-off between bias and variance. We choose $L=T$, which is found to achieve good performance across all tested RL tasks. 

\textbf{Target network update: }
Another potential problem of CARSM is the dependency between the action-value function and policy. Though CARSM is a policy-gradient based algorithm, the gradient estimation procedure is closely related with the action-value function, which may lead to divergence of the estimation as mentioned in previous studies \citep{mnih2016asynchronous,lillicrap2015continuous,bhatnagar2009convergent,maei2010toward}. Fortunately, this issue has been addressed, to some extent, with the help of target network update \citep{mnih2013playing,lillicrap2015continuous}, and we borrow that idea into CARSM for computing policy gradient. In detail, we construct two target networks corresponding to the policy network and $Q$ critic network, respectively; when computing the target of critic network in~\eqref{Q_target}, instead of using the current policy network and $Q$ critic, we use a smoothed version of them to obtain the target value, which can be expressed as
\begin{equation*}
y_{\ell}^{\text{off}} = r(\sv_{\ell},\av_{\ell}) + \gamma\mathbb{E}_{\tilde{\av}\sim\pi'(\cdotv \given \sv_{\ell}')} Q'_{\omegav}(\sv'_{\ell}, \tilde{\av}),
\end{equation*}
where $\pi'$ and $Q'_{\omegav}$ denote the target networks. These target networks are updated every episode in a ``soft'' update manner, as in \citet{lillicrap2015continuous}, by
\begin{equation*}
\begin{aligned}
\omegav^{Q'} \leftarrow \tau \omegav^{Q} + (1-\tau)\omegav^{Q'},~~
\thetav^{\pi'} \leftarrow \tau \thetav^{\pi} + (1-\tau)\thetav^{\pi'},
\end{aligned}
\end{equation*}
which is an exponential moving average of the policy network and action value function network parameters, with $\tau$ as the smoothing parameter. 

\textbf{Annealing on entropy term}:
In practice, maximizing the maximum entropy (ME) objective with an annealing coefficient is often a good choice to encourage exploration and achieve a better sub-optimal solution, and the CARSM gradient estimator for ME would be
\begin{equation*}
\textstyle \gv^{\text{ME}}_{\text{CARSM}} = \gv_{\text{CARSM}} +\lambda\sum_{k=1}^K \nabla_{\thetav}\mathbb{H}(\pi_{\thetav}(a_t|\sv_t)),
\end{equation*}
where $\mathbb{H}(\cdot)$ denotes the entropy term and $\lambda$ is the annealing coefficient. The entropy term can be expressed explicitly because $\pi$ is factorized over its dimensions and there are finite actions along each dimension.

\textbf{Delayed update}: As an accurate critic plays an important role for ARSM to estimate gradient, it would be helpful to adopt the delayed update trick of \citet{fujimoto2018addressing}. In practice, we update the critic network several times before updating the policy network. 

In addition to the Python (TensorFlow) code in the Supplementary Material, we also provide detailed pseudo code to help understand the implementation of CARSM in Appendix \ref{pseudo_code}.

\section{\uppercase {Experiments}}
Our experiments aim to answer the following questions: 
\textbf{(a)} How does the proposed CARSM algorithm perform when compared with ARSM-MC (when ARSM-MC is not too expensive to run)? \textbf{(b)} Is CARSM able to efficiently solve tasks with a large discrete action space? \textbf{(c)} Does CARSM have better sample efficiency than the algorithms, such as A2C and RELAX, that have the same idea of using baselines for variance reduction? \textbf{(d)} Can CARSM be integrated into %
more sophisticated RL learning frameworks %
such as TRPO to achieve an improved performance? Since we run trials on some discretized continuous control tasks, another fair question would be: \textbf{(e)} Will discretization help learning? If so, what are possible explanations?

We %
consider benchmark tasks provided by OpenAI Gym classic-control and MuJoCo %
simulators \citep{todorov2012mujoco}.
We compare the proposed CARSM with ARSM-MC \citep{ARSM}, A2C \citep{mnih2016asynchronous}, and RELAX \citep{grathwohl2017backpropagation}; all of them rely on introducing baseline functions to reduce gradient variance, making it fair to compare them against each other. We then integrate CARSM into TRPO by replacing its A2C gradient estimator for $\nabla_{\thetav}J(\thetav)$. Performance evaluation %
show that a simple plug-in of CARSM estimator can bring the improvement. Details on experimental settings can be found in Appendix \ref{exp_set}. 
 \begin{figure*}[t]
 \includegraphics[width=1\textwidth,height=3.45cm]
 {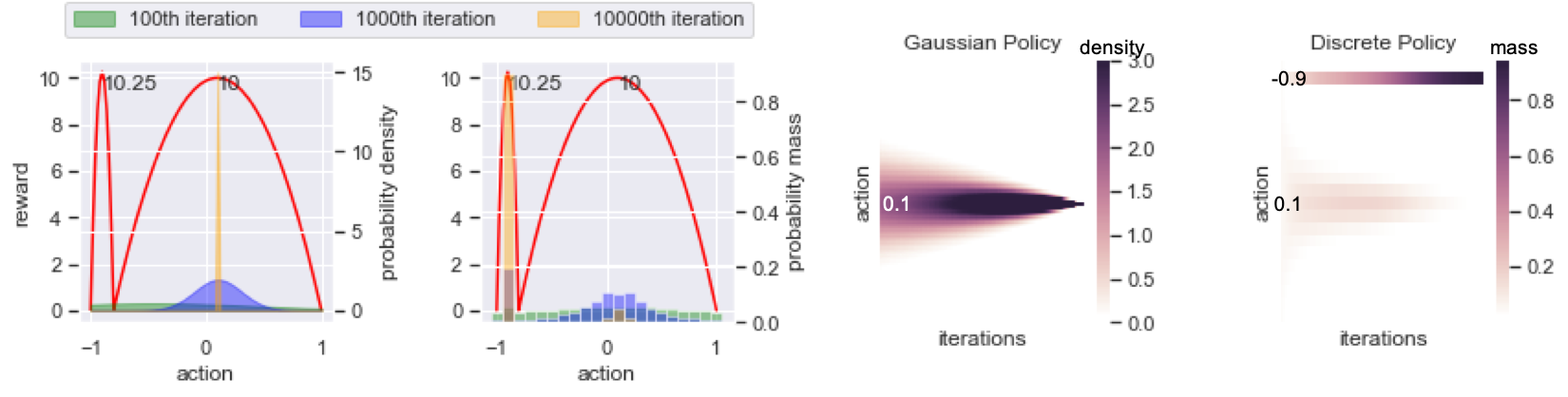}\vspace{-3mm}
\caption{\textbf{left panel}: Change of policy over iterations in a single random trial between Gaussian policy (left) and discrete policy (right) on a bimodal-reward toy example. \textbf{right panel}: Average density on each action along with the training iteration between Gaussian and discrete policies for $100$ random trials. Under this setting, the Gaussian policy fails to converge to the global optimum while discrete policy always finds the global optimum.
}\vspace{-3mm}
 \label{fig:toy_new} 
\end{figure*}

On our experiments with tasks in continuous control domain, we discretize the continuous action space uniformly to get a discrete action space. More specifically, if the action space is $\mathcal{A}=[-1,1]^K$, and we discretize it to $C$ actions at each dimension, the action space would become $\tilde{\mathcal{A}} =
\{\frac{-C+1}{C-1},\frac{-C+3}{C-1},\ldots,\frac{C-1}{C-1}\}^K
$.

There are two motivations of discretizing the action space. First, MuJoCo tasks are a set of standard comparable tasks that naturally have multidimensional action spaces, which is the case we are interested in for CARSM. Second, as illustrated in \citet{tang2019discretizing}, discrete policy is often more expressive than diagonal-Gaussian policy, leading to better exploration. We will illustrate this point by experiments. %

\begin{figure*}[t]
 \centering
 \includegraphics[width=.825\textwidth,height=5.6cm]{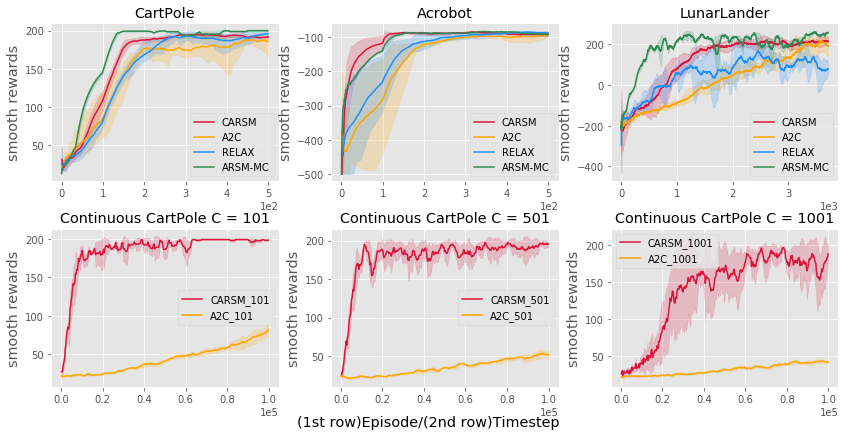}\vspace{-2mm}
\caption{\textbf{top row}: Performance curves for discrete domains. Comparison between: A2C, RELAX, ARSM-MC, and CARSM. We show the cumulative rewards during training, moving averaged across $100$ epochs; the curves show the $\text{mean} \pm \text{std}$ performance across $5$ random seeds. \textbf{bottom row}: Performance curves on CartPole with very large discrete action space. Comparison between: A2C and CARSM over a range of different discretization scale $C\in\{101,501,1001\}$. We show the cumulative rewards during training, moving averaged across $100$ epochs; the curves show the $\text{mean} \pm \text{std}$ performance across $5$ random seeds. }\vspace{-3mm}
 \label{fig:largeC} 
\end{figure*}
\subsection{Motivation and Illustration}\label{motivation}
One distinction between discrete and Gaussian policies is that a discrete policy can %
learn multi-modal and skewed distributions while a Gaussian policy can only support uni-modal, symmetric, and bell-shaped distributions. %
This intrinsic difference
could lead to significantly difference on exploration, as reflected by the toy example presented below, %
which 
will often lead to different sub-optimal solutions in practice. %

To help better understand the connections between multi-modal policy and exploration, we take a brief review of RL objective function from an energy-based distribution point of view. For a bandit problem with reward function $r(a): \mathcal{A}\rightarrow \mathbb{R}$, we denote the true reward induced distribution as
$ %
 p(a)\propto e^{r(a)}.
$ %
The objective function in (\ref{obj}) can be reformulated as %
\begin{equation*}
 \E_{a\sim\pi_{\theta}(a)}[ r(a)] = -\mathbb{KL}(\pi_{\theta}(a)||p(a)) - \mathbb{H}(\pi_{\theta}).
\end{equation*}
The KL-divergence term matches the objective function of variational inference (VI) \citep{blei2017variational} in approximating $p(a)$ with distribution $\pi_{\theta}(a)$, while the second term is the entropy of policy $\pi_{\theta}$. Therefore, if we use maximum entropy objective \citep{haarnoja2017reinforcement}, which is maximizing $\E_{a\sim\pi_{\theta}(a)} [r(a)] +\mathbb{H}(\pi_{\theta})$, we will get an VI approximate solution. %
Suppose $p(a)$ is a multi-modal distribution,
due to the inherent property of VI \citep{blei2017variational}, 
if $\pi_{\theta}$ is a Guassian distribution, it will often underestimate the variance of $p(a)$ and capture only one density mode. By contrast, if $\pi_{\theta}$ is a discrete distribution, it can capture the multi-modal property of $p(a)$, which will lead to more exploration before converging to a more deterministic policy.

We design a simple toy example to reflect these differences. We restrict the action space to $[-1,1]$, and the true reward function is a concatenation of two quadratic functions (as shown in Figure \ref{fig:toy_new} left panel red curves) that intersect at a middle point $m$. We fix the left sub-optimal point as the global optimal one and control the position of $m$ to get tasks with various difficulty levels. More specifically, the closer $m$ to $-1$, the more explorations needed to converge to the global optimal. We defer the detailed experiment setting to Appendix \ref{toy_setup}.
We run $100$ trials of both Gaussian policy and discrete policy and show their behaviors.

 \begin{figure*}[th]
 \centering
 \includegraphics[width=.8\textwidth,height=6.4cm]{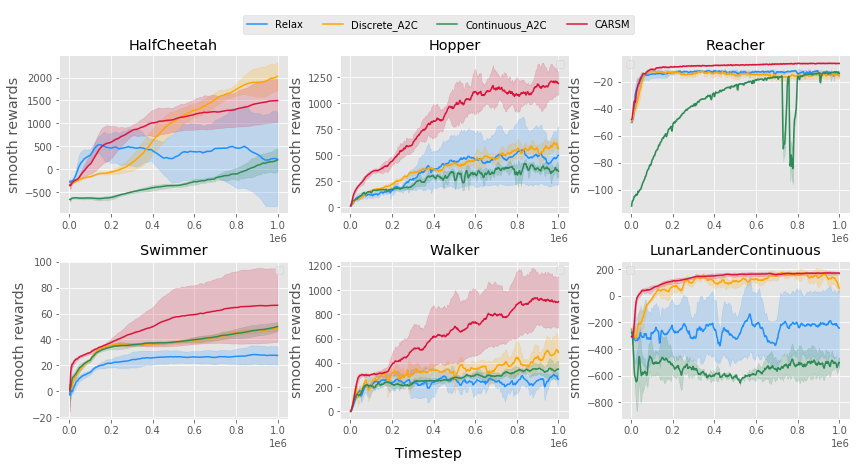}\vspace{-3mm}
\caption{ Performance curves on six benchmark tasks (all except the last are MuJoCo tasks). Comparison between: continuous A2C (Gaussian policy), discrete A2C, RELAX, and CARSM policy gradient. We show the cumulative rewards during training, moving averaged across $100$ epochs; The curves show the $\text{mean} \pm \text{std}$ performance across $5$ random seeds.}
 \label{fig:arsm} \vspace{-3mm}
\end{figure*}

We show, in Figure \ref{fig:toy_new} left panel,  the learning process of both Gaussian and discrete policies with a quadratic annealing coefficient for the entropy term, and, in right panel,
a heatmap where each entry indicates the average density of each action at one iteration.
In this case where $m = -0.8$, %
the signal from the global optimal point has a limited range which requires more explorations during the training process. Gaussian policy can only explore with unimodal distribution and fail to capture the global optimal all the time. By contrast, discrete policy can learn the bi-modal distribution in the early stage, gradually concentrate on both the optimal and sub-optimal peaks before collecting enough samples, and eventually converges to the 
optimal peak. More explanations can be found in Appendix \ref{toy_setup}.

\subsection{Comparing CARSM and ARSM-MC}
One major difference between CARSM and ARSM-MC is the usage of $Q$-Critic. It saves us from running MC rollouts to estimate the action-value functions of all unique pseudo actions, the number of which can be enormous under a multidimensional setting. This saving is at the expense of introducing bias to gradient estimation (not by the gradient estimator per se but by how $Q$ is estimated).
Similar to the argument between MC %
and TD, %
there is a trade-off between bias and variance. In this set of experiments, we show that the use of Critic in CARSM not only brings us accelerated training, but also helps return good performance.

To make the results of CARSM directly comparable with those of ARSM-MC shown in \citet{ARSM}, we evaluate the performances on an \textit{Episode} basis on discrete classical-control tasks: CartPole, Acrobot, and LunarLander. 
We follow \citet{ARSM} to limit the MC rollout sizes for ARSM-MC as $16$, $64$, and $1024$, respectively. 
From Figure \ref{fig:largeC} top row, ARSM-MC has a better performance than CARSM on both CartPole and LunarLander, while CARSM outperforms the rest on Acrobot. The results are promising in the sense that CARSM only uses one rollout for estimation while ARSM-MC uses up to $16$, $64$, and $1024$, respectively, 
so CARSM largely improves the sample efficiency of ARSM-MC while maintaining comparable performance. {The action space is uni-dimensional with $2$, $3$, and $4$ discrete actions for CartPole, Acrobot, and LunarLander, respectively.}
{We also compare  CARSM and ARSM-MC given fixed number of timesteps. Under this setting, CARSM outperforms ARSM-MC by a large margin on both Acrobot and LunarLander. See Figure~\ref{fig_appendix} in Appendix~\ref{comp_fix} for more details.}

\begin{figure*}[th]
 \centering
 \includegraphics[width=.8\textwidth,height=6.4cm]{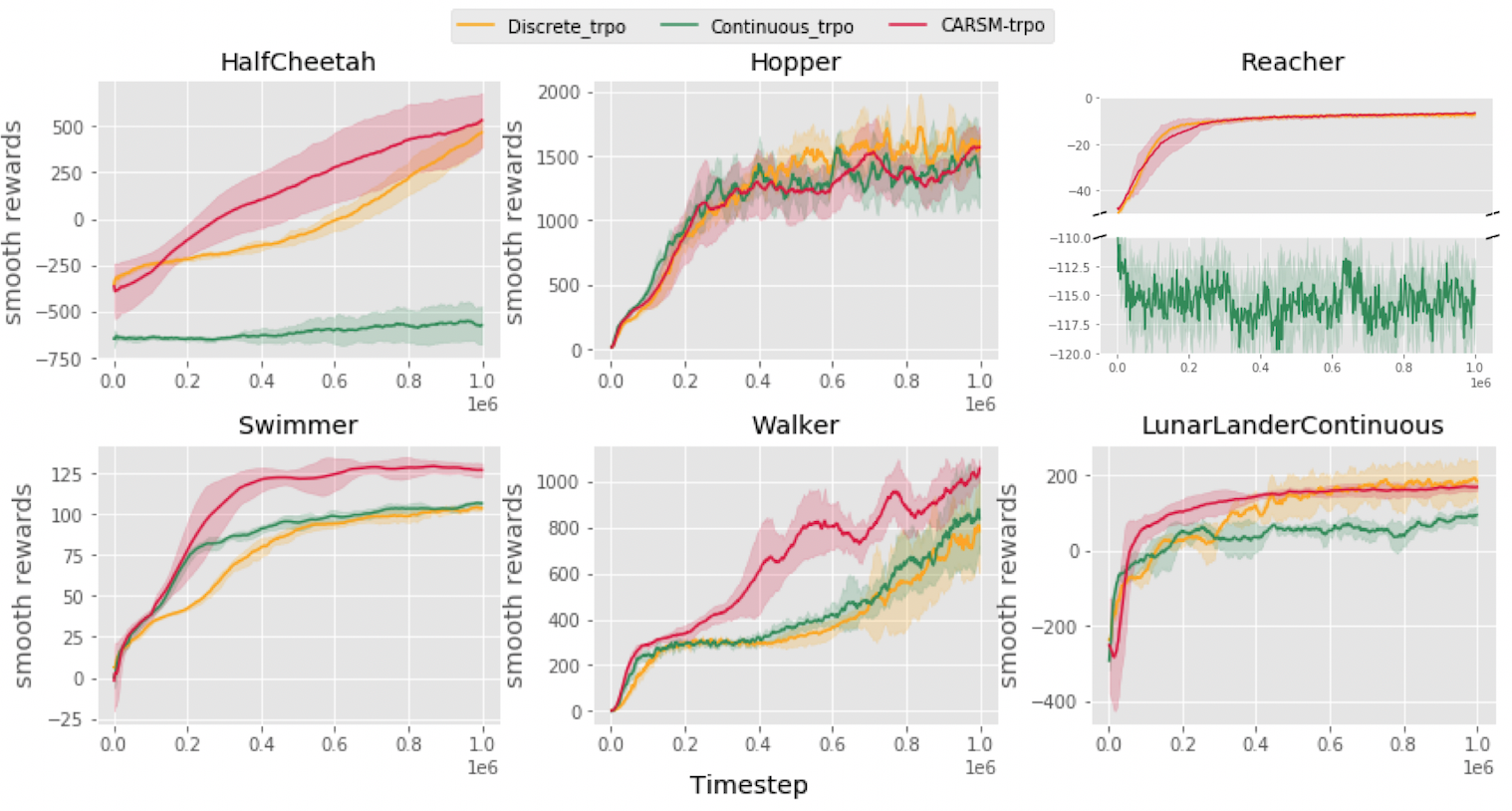}\vspace{-3mm}
\caption{ Performance curves on six benchmark tasks (all except the last are MuJoCo tasks). Comparison between: continuous TRPO (Gaussian policy), discrete TRPO, and CARSM policy gradient combined with TRPO. We show the cumulative rewards during training, moving averaged across $100$ epochs; the curves show the $\text{mean} \pm \text{std}$ performance across $5$ random seeds. }\vspace{-3mm}
 \label{fig:trpo} 
\end{figure*}

\begin{figure*}[th]
 \centering
 \begin{subfigure}[t]{0.18\textwidth}
 \centering
 \includegraphics[width=\textwidth]{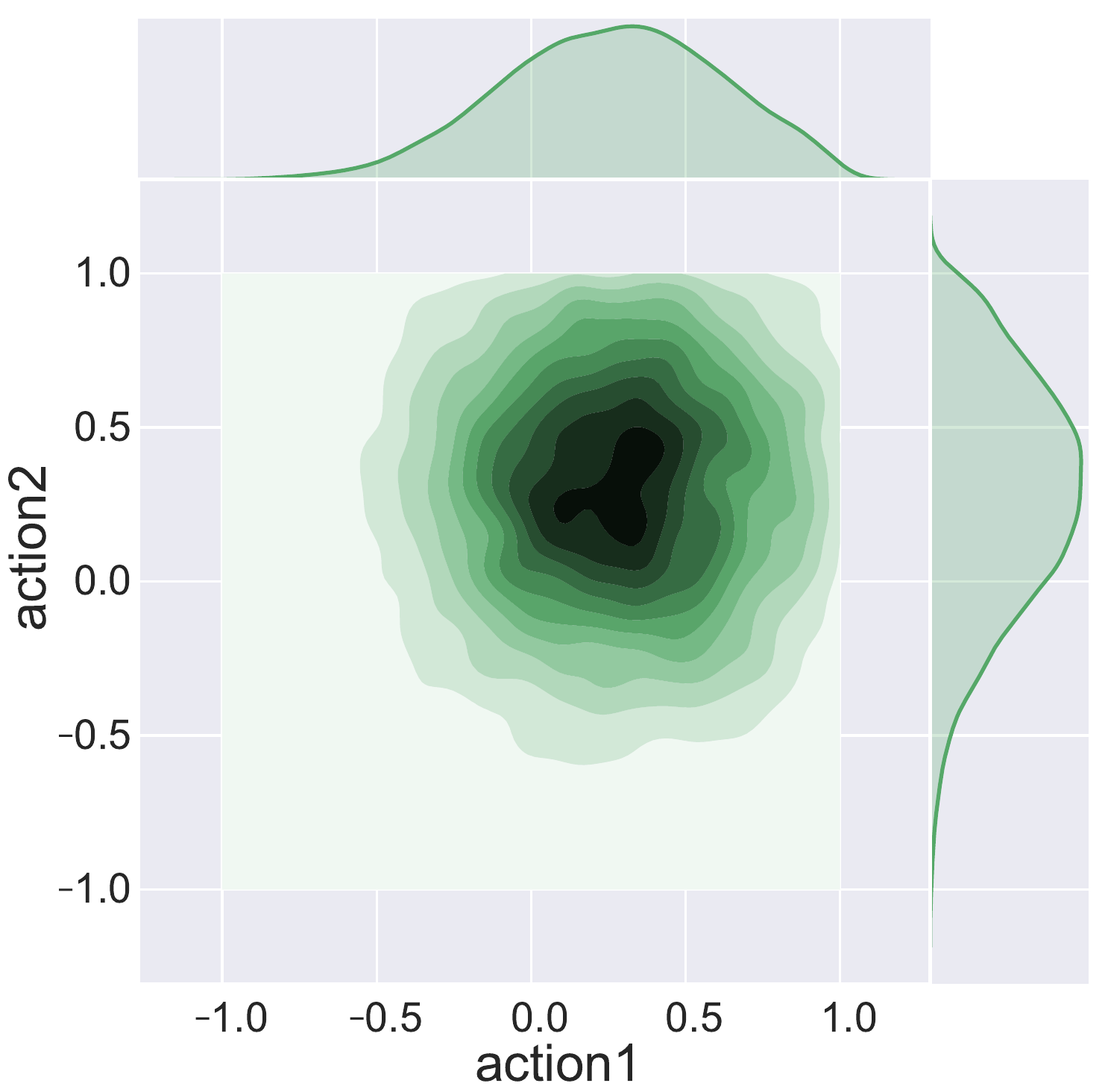}
 \caption[]%
 {{\small Gaussian policy at an early stage.} } 
 \end{subfigure}
 \hfill
 \begin{subfigure}[t]{0.18\textwidth} 
 \centering 
 \includegraphics[width=\textwidth]{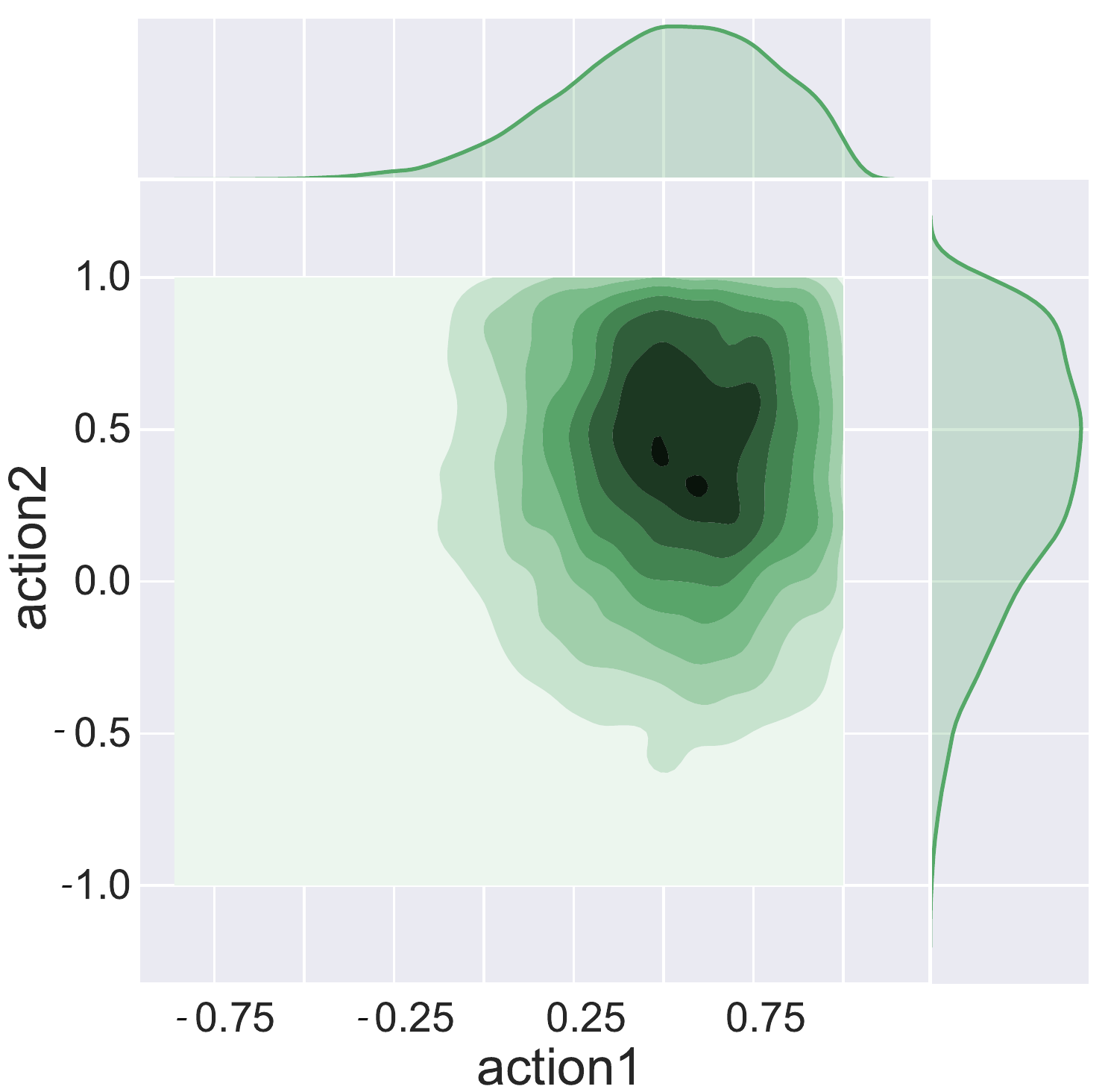}
 \caption[]%
 {{\small Gaussian policy at an intermediate stage.}}
 \end{subfigure}
 \hfill
 \begin{subfigure}[t]{0.18\textwidth} 
 \centering 
 \includegraphics[width=\textwidth]{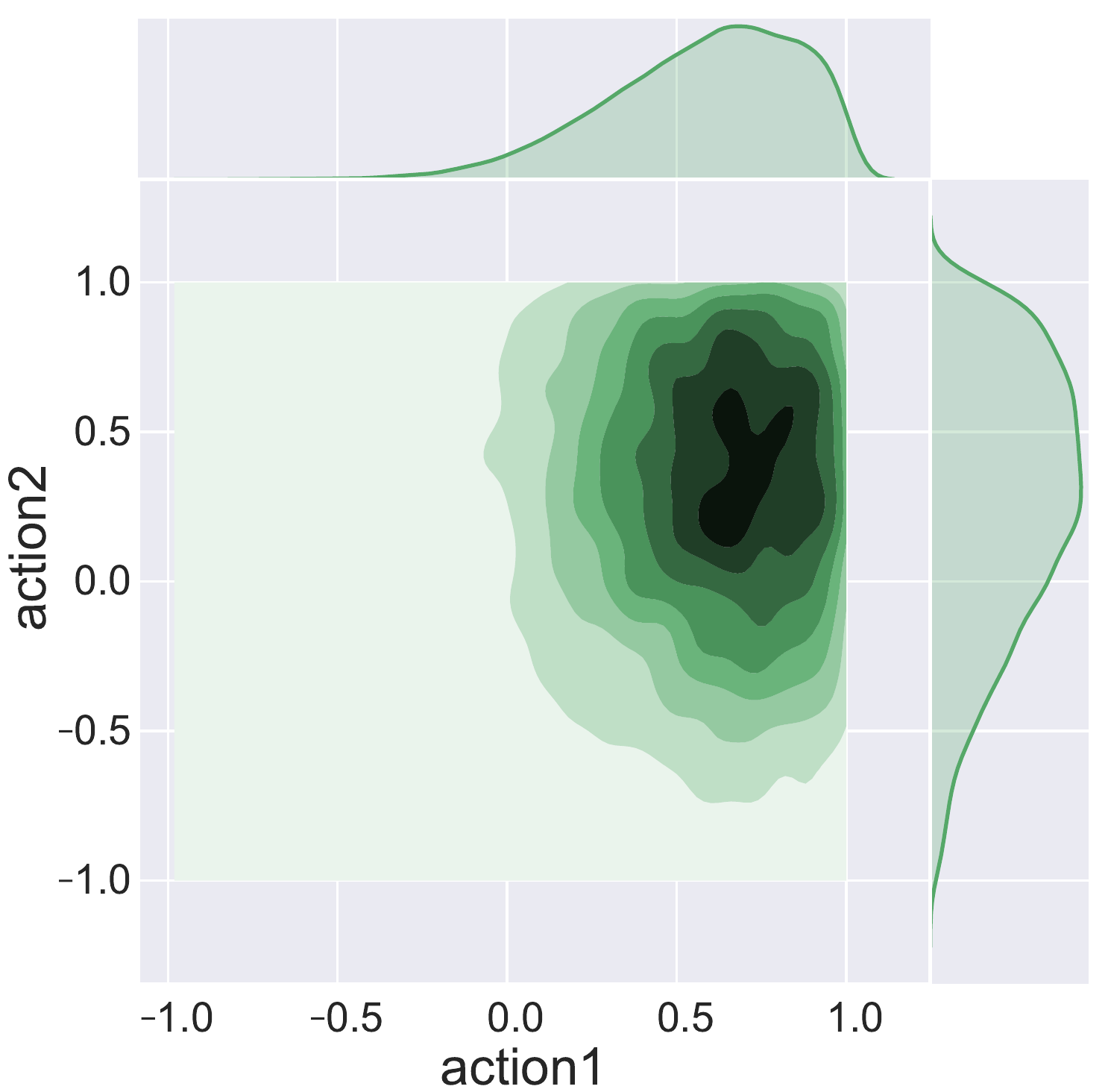}
 \caption[]%
 {{\small Gaussian policy at a late stage.}}
 \end{subfigure}
 \hfill
 \begin{subfigure}[t]{0.18\textwidth} 
 \centering 
 \includegraphics[width=\textwidth]{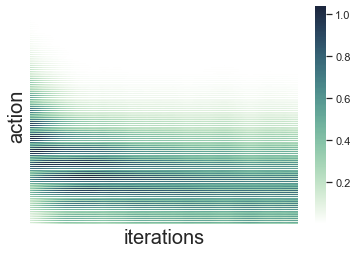}
 \caption[]%
 {{\small Average density for Gaussian policy.}} 
 \end{subfigure}
 \\
 \begin{subfigure}[t]{0.18\textwidth}
 \centering
 \includegraphics[width=\textwidth]{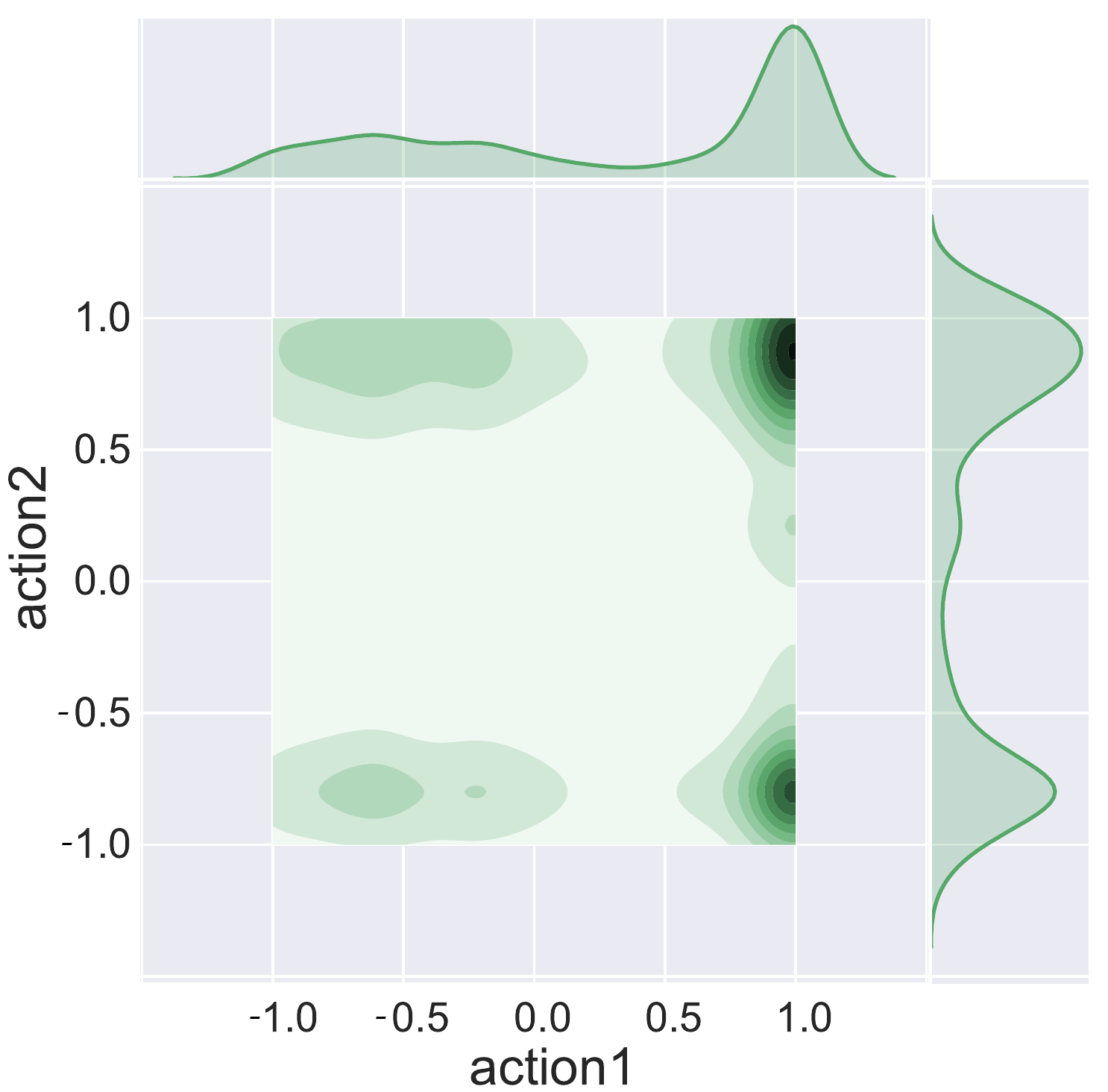}
 \caption[]%
 {{\small Discrete policy at an early stage.}} 
 \label{fig:cont_1}
 \end{subfigure}
 \hfill
 \begin{subfigure}[t]{0.18\textwidth} 
 \centering 
 \includegraphics[width=\textwidth]{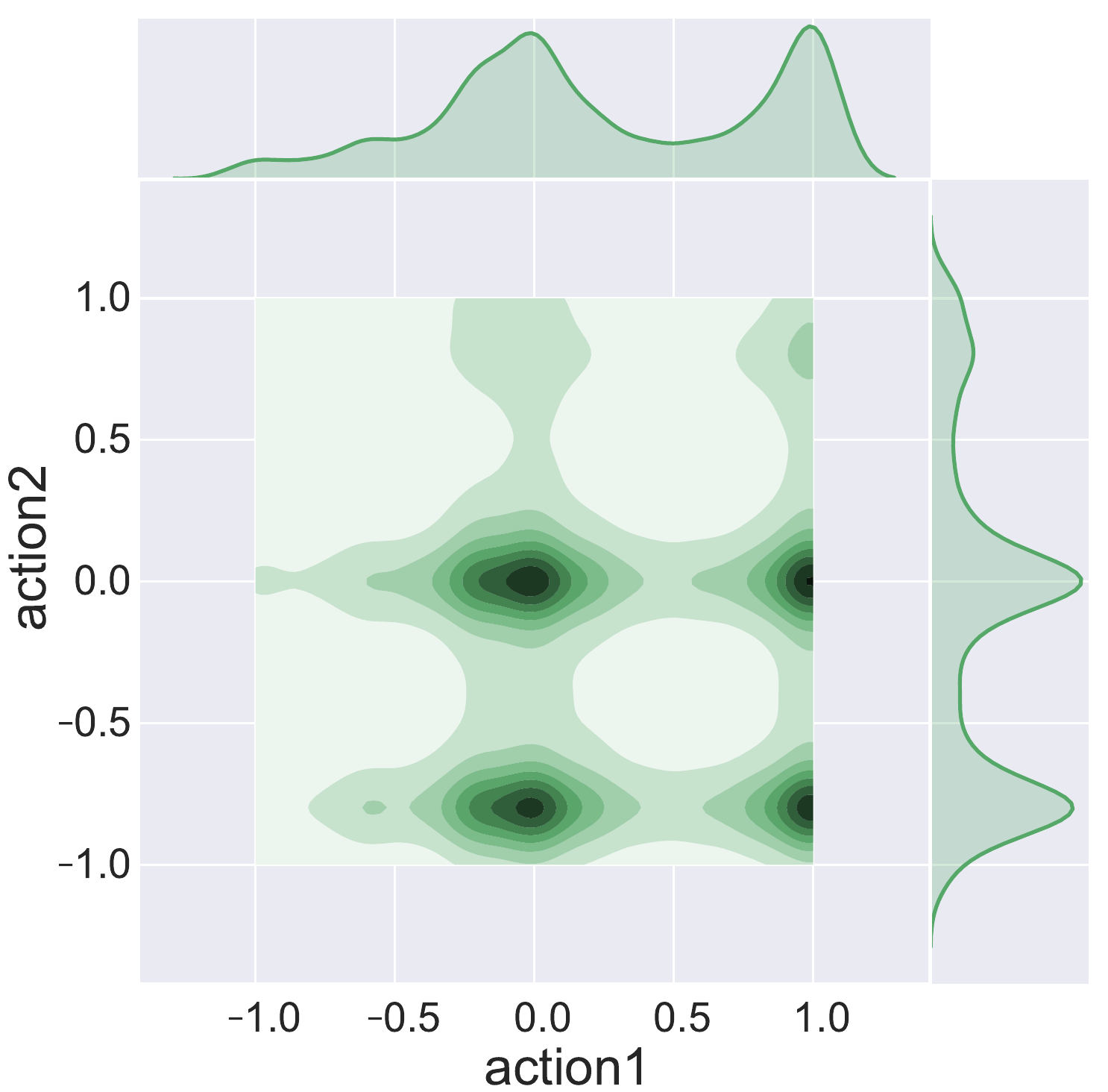}
 \caption[]%
 {{\small Discrete policy at an intermediate stage.}}
 \label{fig:cont_2}
 \end{subfigure}
 \hfill
 \begin{subfigure}[t]{0.18\textwidth} 
 \centering 
 \includegraphics[width=\textwidth]{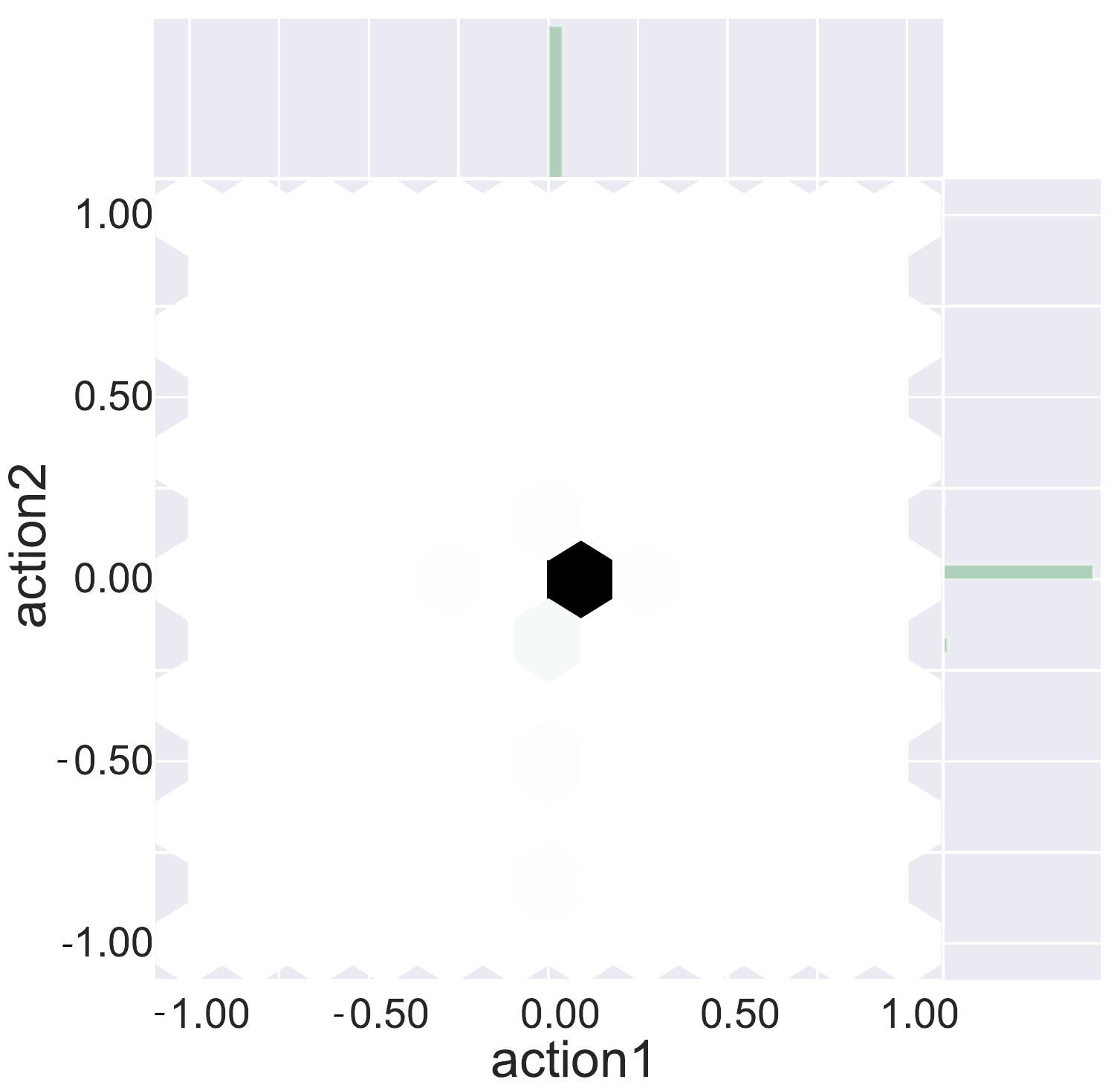}
 \caption[]%
 {{\small Discrete policy at a late stage.}}
 \label{fig:cont_3}
 \end{subfigure}
 \hfill
 \begin{subfigure}[t]{0.18\textwidth} 
 \centering 
 \includegraphics[width=\textwidth]{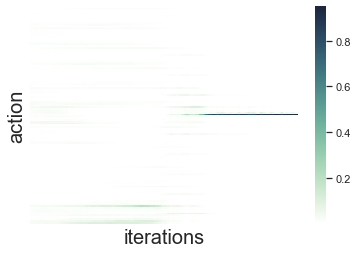}
 \caption[]%
 {{\small Average density for Discrete policy.}} 
 \label{fig:cont_4}
 \end{subfigure}
   \vspace{-2mm}
 \caption[]
 {\small Policy distribution on the Reacher task between discrete policy and Gaussian policy for a given state (discrete action space has $11$ actions on each dimension). 
} 
 \label{fig:reacher_evolve}\vspace{-4mm}
 \end{figure*}

\subsection{Large Discrete Action Space}
We want to show that CARSM has better sample efficiency on cases where the number of action $C$ in one dimension is large. We test CARSM along with A2C on a continuous CartPole task, which is a modified version of discrete CartPole. In this continuous environment, we restrict the action space to $[-1,1]$. Here the action indicates the force applied to the Cart
at any time. 

The intuition of why CARSM is expected to perform well under a large action space setting is because of the low-variance property. When $C$ is large, the distribution is more dispersed on each action compared with smaller case, which requires the algorithm captures the signal from best action accurately to improve exploitation. In this case, a high-variance gradient estimator will surpass the right signal, leading to a long exploration period or even divergence. %

As shown in Figure \ref{fig:largeC} bottom row, CARSM outperforms A2C by a large margin in all three large $C$ settings. %
Though the CARSM curve exhibits larger variations as $C$ increases, it always learns
much more rapidly at an early stage compared with A2C. Note the naive ARSM-MC algorithm will not work on this setting simply because it needs to run as many as tens of thousands MC rollouts to get a single gradient estimate. 
%

\subsection{OpenAI Gym Benchmark Tasks}\label{mujoco_benchmark}
In this set of experiments, we compare CARSM with A2C and RELAX, which all share the same underlying idea of improving the sample efficiency by reducing the variance of gradient estimation. For A2C, we compare with both Gaussian and discrete policies to check the intuition  presented in Section \ref{motivation}.
In all these tasks, following the results from \citet{tang2019discretizing}, the action space is equally divided into $C=11$ discrete actions at each dimension. %
Thus the discrete action space size becomes $11^K$, where $K$ is the action-space dimension that is $6$ for HalfCheetah, $3$ Hopper, $2$ Reacher, $2$ Swimmer, $6$ Walker2D, and $2$ LunarLander. More details on Appendix~\ref{env_setup}.

As shown in Figure \ref{fig:arsm}, CARSM outperforms the other algorithms by a large margin except on HalfCheetah,  demonstrating the high-sample efficiency of CARSM. 
Moreover, the distinct behaviors of Gaussian and discrete policies in the Reacher task, as shown in both Figures \ref{fig:arsm} and \ref{fig:trpo}, are worth thinking, motivating us to go deeper on this task to search for possible explanations. We manually select a state that requires exploration on the early stage, and record the policy evolvement along with training process at that specific state. We show those transition phases in Figure \ref{fig:reacher_evolve} for both Gaussian policy (top row) and discrete policy (bottom row). 

For discrete policy, plots (e)-(g) in Figure \ref{fig:reacher_evolve} bottom row show interesting property: at the early stage, the policy does not put heavy mass at all on the final sub-optimal point $(0,0)$, but explores around multiple density modes; then it gradually concentrates on several sup-optimal points on an intermediate phase, and converges to the final sub-optimal point. Plot (h) also conveys the same message that during the training process, discrete action can transit explorations around several density modes since the green lines can jump along the iterations. (The heatmaps of (d) and (h) in Figure \ref{fig:reacher_evolve} are computed in the same way as that in Figure \ref{fig:toy_new}, and details can be found in Appendix \ref{toy_setup}.)

By contrast, Gaussian policy does not have the flexibility of exploring based on different density modes, therefore from plots (a)-(c) on the top row of Figure~\ref{fig:reacher_evolve}, the policy moves with a large radius but one center, and on (d), the green lines move consecutively which indicates a smooth but potentially not comprehensive exploration.

\subsection{Combining CARSM with TRPO}
Below we show that CARSM can be readily applied under TRPO to improves its performance. In the update step of TRPO shown in (\ref{trpo}), the default estimator for $\nabla_{\thetav}J(\thetav)$ is A2C or its variant. We replace it with CARSM estimator and run it on the same set of tasks. 
As shown in Figure \ref{fig:trpo}, 
Gaussian policy fails to find a good sub-optimal solution under TRPO for both HalfCheetah and Reacher and performs similarly to its discrete counterpart on the other tasks. Meanwhile,
CARSM improves the performance of TRPO over discrete policy setting on three tasks and maintains similar performance on the others, which shows evidence that CARSM is an easy plug-in estimator for $\nabla_{\thetav}J(\thetav)$ and hence can potentially improve other algorithms, such as some off-policy ones \citep{wang2016sample,degris2012off}, that need this gradient estimation. 

\section{\uppercase {Conclusion}}
To solve RL tasks with multidimensional discrete action setting efficiently, we propose Critic-ARSM policy gradient, which is a combination of multidimensional sparse ARSM gradient estimator and an action-value critic, to improve sample efficiency for on-policy algorithm. 
We show the good performances of this algorithm from perspectives including stability on very large action space cases and comparisons with other standard benchmark algorithms, and show its potential to be combined with other standard algorithms. Moreover, we demonstrate the potential benefits of discretizing continuous control tasks to obtain a better exploration based on multimodal property. 

\newpage
\subsection*{Acknowledgements}
This research was supported in part by Award IIS-1812699
from the U.S. National Science Foundation and the 2018-2019 McCombs Research Excellence Grant. The authors acknowledge the support
of NVIDIA Corporation with the donation of the Titan Xp GPU used for this research, and the
computational support of Texas Advanced Computing Center.
\balance{
\bibliographystyle{abbrvnat}
\bibliography{reference.bib}
}
\newpage
\appendix
\onecolumn
\begin{center}{\large{\textbf{Discrete Action On-Policy Learning with Action-Value Critic:\\ \vspace{2mm} Supplementary Material}}}\end{center}

\section{Proof of Theorem \ref{thm1}}\label{proof_thm1}
We first show the sparse ARSM for multidimensional action space case at one specific time point, then generalize it to stochastic setting. 
Since $a_{k}$ are conditionally independent given $\phiv_{k}$, the gradient of $\phiv_{kc}$ at one time point would be (we omit the subscript $t$ for simplicity here)
\bas{
\nabla_{ \phi_{kc}}J(\phiv)&=\E_{\av_{\backslash k}\sim\prod_{k'\neq k}{\text{Discrete}}(a_{k'};\sigma(\phiv_{k'}))} \big[ \nabla_{ \phi_{kc}} \E_{a_{k}\sim {\text{Cat}}(\sigma(\phiv_{k}))} [Q{}(\av,\sv)]\big ],
}
and we apply the ARSM gradient estimator on the inner expectation part, which gives us
\ba{
\small
\small\nabla_{\phi_{kc}}J(\phiv)&\small =\E_{\av_{\backslash k}\sim\prod_{k'\neq k}{\text{Discrete}}(a_{k'};\sigma(\phiv_{k'}))} \Big\{ \E_{ \varpiv_{k}\sim{\text{Dir}}(\mathbf{1}_{C})}\Big[(Q([\av_{\backslash k},a_k^{c \leftrightharpoons j}],\sv )-\frac{1}{C} \sum_{m=1}^C Q([\av_{\backslash k},a_k^{m \leftrightharpoons j}],\sv ))(1-C \varpi_{kj})\Big]\Big\}\notag\\
&\small = \E_{ \varpiv_{k}\sim{\text{Dir}}(\mathbf{1}_{C})} \Big\{ \E_{\av_{\backslash k}\sim\prod_{k'\neq k}{\text{Discrete}}(a_{k'};\sigma(\phiv_{k'}))}\Big[(Q([\av_{\backslash k},a_k^{c \leftrightharpoons j}],\sv )-\frac{1}{C} \sum_{m=1}^C Q([\av_{\backslash k},a_k^{m \leftrightharpoons j}],\sv ))(1-C \varpi_{kj})\Big]\Big\}
\label{eq:PHI_mv_0}
\\
&\small = \E_{ \varpiv_{k}\sim{\text{Dir}}(\mathbf{1}_{C})} \Big\{ \E_{\prod_{k'\neq k}{\text{Dir}}(\varpiv_{k'};\mathbf{1}_C)}\Big[(Q(\av^{c \leftrightharpoons j},\sv )-\frac{1}{C} \sum_{m=1}^C Q(\av^{m \leftrightharpoons j},\sv ))(1-C \varpi_{kj})\Big]\Big\}\label{eq:PHI_mv_1},
} 
where \eqref{eq:PHI_mv_0} is derived by changing the order of two expectations
and \eqref{eq:PHI_mv_1} can be derived by following the proof of Proposition 5 in \citet{ARSM}.
Therefore, if given $\varpiv_k\sim\mbox{Dir}(\mathbf 1_C)$, it is true that $a_{k}^{_{{c} \leftrightharpoons j}} = a_{k}$ for all $(c,j)$ pairs, then the inner expectation term in \eqref{eq:PHI_mv_0} will be zero and consequently we have
\begin{equation*}
g_{kc} = 0
\end{equation*}
as an unbiased single sample estimate of $\nabla_{\phi_{kc}}J(\phiv)$; %
If given $\varpiv_k\sim\mbox{Dir}(\mathbf 1_C)$, there exist $(c,j)$ that $a_{k}^{_{{c} \leftrightharpoons j}} \neq a_{k}$, we 
can 
use \eqref{eq:PHI_mv_1} to provide 
\ba{
g_{kc}= \sum_{j=1}^C \left[Q(\sv,\av^{_{{c} \leftrightharpoons j}} ) - \frac{1}{C}\sum_{m=1}^{C} Q(\sv,\av^{_{{m} \leftrightharpoons j}} )\right] \left(\frac{1}{C}-\varpi_{kj}\right) %
 \label{arsm_multi}
 }
 as an unbiased single sample estimate of $\nabla_{\phi_{kc}}J(\phiv)$.
 
For a specific time point $t$, the objective function can be decomposed as
\bas{
\textstyle 
J(\phiv_{0:\infty})&=\mathbb{E}_{\mathcal{P}(\sv_{0})\left[\prod_{t'=0}^{t-1}\mathcal{P}(\sv_{t'+1}\given \sv_{t'},\av_{t'}) \text{Cat}(\av_{t'};\sigma(\phiv_{t'}))\right]}\left\{\E_{\av_{t}\sim \text{Cat}(\sigma(\phiv_{t})) }\left[\sum_{t'=0}^{t-1}\gamma^{t'}r(\sv_{t'},\av_{t'})+ \gamma^{t}Q(\sv_t,\av_t)\right]\right\}\notag\\
&=\mathbb{E}_{\mathcal{P}(\sv_{0})\left[\prod_{t'=0}^{t-1}\mathcal{P}(\sv_{t'+1}\given \sv_{t'},\av_{t'}) \text{Cat}(\av_{t'};\sigma(\phiv_{t'}))\right]}\left\{\E_{\av_{t}\sim \text{Cat}(\sigma(\phiv_{t})) }\left[\sum_{t'=0}^{t-1}\gamma^{t'}r(\sv_{t'},\av_{t'})\right]\right\}\notag\\
&~~~~+\mathbb{E}_{\mathcal{P}(\sv_{0})\left[\prod_{t'=0}^{t-1}\mathcal{P}(\sv_{t'+1}\given \sv_{t'},\av_{t'}) \text{Cat}(\av_{t'};\sigma(\phiv_{t'}))\right]}\left\{\E_{\av_{t}\sim \text{Cat}(\sigma(\phiv_{t})) }\left[ \gamma^{t}Q(\sv_t,\av_t)\right]\right\},
}
where the first part has nothing to do with $\phiv_{t}$, we therefore have
\bas{
\nabla_{\phiv_{tkc}}J(\phiv_{0:\infty})& =
\E_{\mathcal{P}(\sv_{t}\given \sv_0,\pi_{\thetav}) \mathcal{P}(\sv_{0}) }\left\{\gamma^{t} \nabla_{\phiv_{tkc}} \E_{\av_{t}\sim \text{Cat}(\sigma(\phiv_{t})) }\left[ Q(\sv_t,\av_t)\right]\right\}\notag\\.
}
With the result from \eqref{arsm_multi}, the statements in Theorem\ref{thm1} follow. 

\newpage 

\section{Experiment setup}
 \begin{figure*}[t]
 \includegraphics[width=1\textwidth,height=3.45cm]{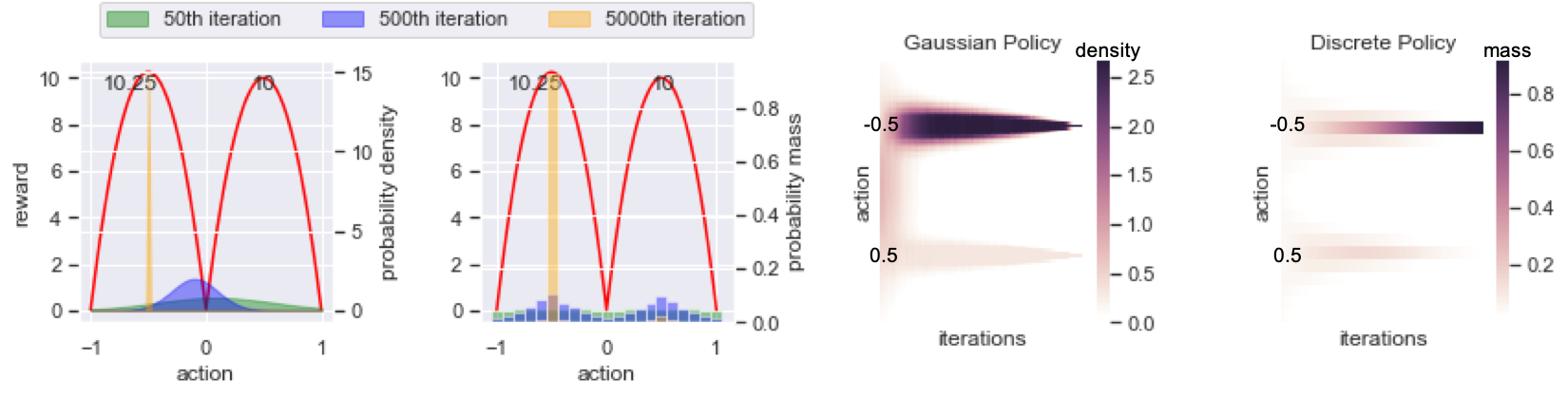}
\caption{\textbf{left panel}: Change of policy over iterations between Gaussian policy (left) and discrete policy (right) on toy example setting. \textbf{right panel}: Average density on each action along with the training iterations between Gaussian policy and discrete policy for $100$ experiments.(The Gaussian policy converges to the inferior optimal solution $12$ times out of $100$ times, and discrete policy converges to the global optimum all the time).}
 \label{fig:toy} 
\end{figure*}
\subsection{Toy example setup}\label{toy_setup}
Assume the true reward is a bi-modal distribution (as shown in Figure \ref{fig:toy} left panel red curves) with a difference between its two peaks:
\begin{equation*}
 r(a)=\left\{
 \begin{array}{ll}
 -c_1 (a-1)(a-m) +\epsilon_1 &\text{for }a\in[m,1]\\
 -c_2(a+1)(a-m)+\epsilon_2 &\text{for } a\in[-1,m],\\
 \end{array}
 \right.
\end{equation*}
where the values of
$c_1$, $c_2$, and $m$ determine the heights and widths of these two peaks, 
and $\epsilon_1\sim$N$(0,2)$ and $\epsilon_2\sim$N$(0,1)$ are noise terms. It is clear that
 $a^*_{\text{left}} =(m-1)/2$ and $a^*_{\text{right}} = (1+m)/2$ are two local-optimal solutions and corresponding to $r_{\text{left}}:=\E [r((a^*_{\text{left}})] = c_2(1+m)^2/4$ and $r_{\text{right}} := \E [r(a^*_{\text{right}})] = c_1(1-m)^2/4$. Here we always choose $c_1$ and $c_2$ such that $r_{\text{left}}$ is slightly bigger than $r_{\text{right}}$ which makes $a^*_{\text{left}}$ a better local-optimal solution. It is clear that the more closer $a^*_{\text{left}}$ to $-1$, the more explorations a policy will need to converge to $a^*_{\text{left}}$. Moreover, the noise terms can give wrong signals and may lead to bad update directions, and exploration will play an essential role in preventing the algorithm from acting too greedily. 
 The results shown on Section \ref{motivation} has $m = -0.8$, $c_1 = 40/(1.8^2)$ and $c_2 = 41/(0.2^2)$, which makes $r_{\text{left}} = 10.25$ and $r_{\text{right}} = 10$.
 We also show a simple example at Figure \ref{fig:toy} with $m = 0$, $c_1 = 40/(0.5^2)$ and $c_2 = 41/(0.5^2)$, which maintains the same peak values.

The experiment setting is as follows: for each episode, we collect $100$ samples and update the corresponding parameters ($[\mu,\sigma]$ for Gaussian policy and $\phiv\in \mathbb{R}^{21}$ for discrete policy where the action space is discretized to $21$ actions), and iterate until $N$ samples are collected. We add a quadratic decaying coefficient for the entropy term for both policies to encourage explorations on an early stage. 
The Gaussian policy is updated using reparametrization trick \citep{kingma2013auto}, which can be applied to this example since we know the derivative of the reward function (note this is often not the case for RL tasks). The discrete policy is updated using ARSM gradient estimator described in Section \ref{prelim}.

On the heatmap, the horizontal axis is the iterations, and vertical axis denotes the actions. For each entry corresponding to $a$ at iteration $i$, its value is calculated by $v(i,a) = \frac{1}{U}\sum_{u=1}^U p_u(a\given i)$, where $p_u(a\given i)$ is the probability of taking action $a$ at iteration $i$ for that policy in $u$th trial.

We run the same setting with different seeds for Gaussian policy and discrete policy for $100$ times, where the initial parameters for Gaussian Policy is $\mu_0=m, \sigma = 1$ and for discrete policy is $\phi_i = 0$ for any $i$ to eliminate the effects of initialization. 

In those $100$ trials, when $m = -0.8, N = 1e^6$, Gaussian policy fails to find the true global optimal solution ($0/100$) while discrete policy can always find that optimal one ($100/100$). When $m = 0, N = 5e^5$, the setting is easier and 
Gaussian policy performs better in this case with only $12/100$ percentage converging to the inferior sub-optimal point $0.5$, and the rest $88/100$ chances getting to global optimal solution. On the other hand, discrete policy always converges to the global optimum ($100/100$). The similar plots are shown on Figure \ref{fig:toy}. 
The \textit{p-value} for this proportion test is $0.001056$, which shows strong evidence that discrete policy outperforms Gaussian policy on this example.

\subsection{Baselines and CARSM setup}\label{exp_set}
Our experiments aim to answer the following questions: \textbf{(a)} How does the proposed CARSM algorithm perform when compared with ARSM-MC (when ARSM-MC is not too expensive to run). \textbf{(b)} Is CARSM able to efficiently solve tasks with large discrete action spaces (i.e., $C$ is large). \textbf{(c)} Does CARSM have better sample efficiency than the algorithms, such as A2C and RELAX, that have the same idea of using baselines for variance reduction. \textbf{(d)} Can CARSM combined with other standard algorithms such as TRPO to achieve a better performance. 

\paragraph{Baselines and Benchmark Tasks.} We evaluate our algorithm on benchmark tasks on OpenAI Gym classic-control and MuJoCo tasks \citep{todorov2012mujoco}. We compare the proposed CARSM with ARSM-MC \citep{ARSM}, A2C \citep{mnih2016asynchronous}, and RELAX \citep{grathwohl2017backpropagation}; all of them rely on introducing baseline functions to reduce gradient variance, making it fair to compare them against each other. We then integrate CARSM into TRPO by replacing the A2C gradient estimator for $\nabla_{\thetav}J(\thetav)$, and evaluate the performances on MuJoCo tasks to show that a simple plug-in of the CARSM estimator can bring the improvement.

\paragraph{Hyper-parameters: } Here we detail the hyper-parameter settings for all algorithms. Denote $\beta_{\text{policy}}$ and $\beta_{\text{critic}}$ as the learning rates for policy parameters and $Q$ critic parameters, respectively, $n_{\text{critic}}$ as the number of training time for $Q$ critic, and $\alpha$ as the coefficient for entropy term. For CARSM, we select the best learning rates $\beta_{\text{policy}},~\beta_{\text{critic}}\in\{1,3\}\times 10^{-2}$, and $n_{\text{critic}}\in\{50,150\}$; For A2C and RELAX, we select the best learning rates $\beta_{\text{policy}}\in\{3,30\}\times 10^{-5}$. In practice, the loss function consists of a policy loss $L_{\text{policy}}$ and value function loss $L_{\text{value}}$. The policy/value function are optimized jointly by optimizing the aggregate objective at the same time $L = L_{\text{policy}} + c L_{\text{value}}$, where $c = 0.5$. Such joint optimization is popular in practice and might be helpful in cases where policy/value function share parameters. For A2C, we apply a batched optimization procedure: at iteration $t$, we collect data using a previous policy iterate $\pi_{t-1}$. The data is used for the construction of a differentiable loss function $L$. We then take $v_{\text{iter}}$ gradient updates over the loss function objective to update the parameters, arriving at $\pi_t$. In practice, we set $v_{\text{iter}} = 10$. For TRPO and TRPO combined with CARSM, we use max KL-divergence of $0.01$ all the time without tuning. All algorithms use a initial $\alpha$ of $0.01$ and decrease $\alpha$ exponentially, and target network parameter $\tau$ is $0.01$.
To guarantee fair comparison, we only apply the tricks that are related to each algorithm and didn't use any general ones such as normalizing observation. More specifically, we replace Advantage function with normalized Generalized Advantage Estimation (GAE) \citep{schulman2015high} on A2C, apply normalized Advantage on RELAX.

\textbf{Structure of $Q$ critic networks: } There are two common ways to construct a $Q$ network. The first one is to model the network as $Q: \mathbb{R}^{n_S}\rightarrow \mathbb{R}^{|\mathcal{A}|}$, where $n_S$ is the state dimension and $|\mathcal{A}|=C^K$ is the number of unique actions. The other structure is $Q:\mathbb{R}^{n_S+K}\rightarrow \mathbb{R}$, which means we need to concatenate the state vector $\sv$ with action vector $\av$ and feed that into the network. The advantage of first structure is that it doesn't involve the issue that action vector and state vector are different in terms of scale, which may slow down the learning process or make it unstable. However, the first option is not feasible under most multidimensional discrete action situations because the number of actions grow exponentially along with the number of dimension $K$. 
Therefore, we apply the second kind of structure for $Q$ network, and update $Q$ network multiple times before using it to obtain the CARSM estimator to stabilize the learning process. 

\textbf{Structure of policy network}: The policy network will be a function of $\mathcal{T}_{\thetav}: \mathbb{R}^{n_S}\rightarrow \mathbb{R}^{K\times C}$, which feed in state vector $\sv$ and generate $K\times C$ logits $\phi_{kc}$. Then the action is obtained for each dimension~$k$ by $\pi(a_k\given \sv,\thetav) = \sigma(\phiv_k)$, where $\phiv_k=(\phi_{k1},\ldots,\phi_{kC})'$. %
For both the policy and $Q$ critic networks, we use a two-hidden-layer multilayer perceptron %
with $64$ nodes per layer and tanh activation. 

\textbf{Environment setup}\label{env_setup}
\begin{multicols}{2}
\begin{itemize}
    \item \textit{HalfCheetah} ($\mathcal{S}\subset\mathbb{R}^{17}, \mathcal{A}\subset\mathbb{R}^{6}$)
    \item \textit{Hopper} ($\mathcal{S}\subset\mathbb{R}^{11}, \mathcal{A}\subset\mathbb{R}^{3}$)
    \item \textit{Reacher} ($\mathcal{S}\subset\mathbb{R}^{11}, \mathcal{A}\subset\mathbb{R}^{2}$)
    \item \textit{Swimmer} ($\mathcal{S}\subset\mathbb{R}^{8}, \mathcal{A}\subset\mathbb{R}^{2}$)
    \item \textit{Walker2D} ($\mathcal{S}\subset\mathbb{R}^{17}, \mathcal{A}\subset\mathbb{R}^{6}$)   
    \item \textit{LunarLander Continuous} ($\mathcal{S}\subset\mathbb{R}^{8}, \mathcal{A}\subset\mathbb{R}^{2}$)       
\end{itemize}
\end{multicols}

\subsection{Comparison between CARSM and ARSM-MC for fixed timestep}\label{comp_fix}
\begin{figure*}[t]
\centering
 \includegraphics[width=0.7\textwidth,height=3.45cm]{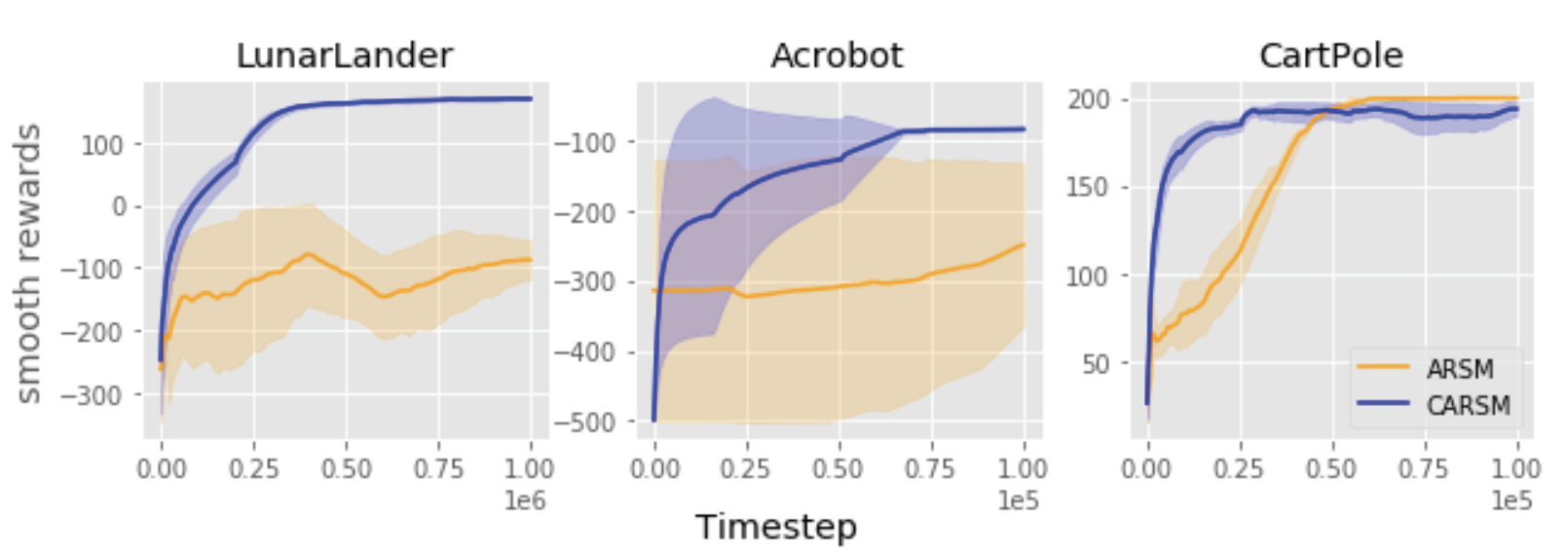}
 \caption{Performance curves for comparison between ARSM-MC and CARSM given fix timesteps}
 \label{fig_appendix}
\end{figure*}

{We compare ARSM-MC and CARSM for fixed timestep setting, with their performances shown in Figure~\ref{fig_appendix}}

\newpage
\section{Pseudo Code}\label{pseudo_code}
We provide detailed pseudo code to help understand the implementation of CARSM policy gradient. There are four major 
steps for each update iteration: (1) Collecting samples using augmented Dirichlet variables $\varpiv_t$; (2) Update the $Q$ critic network using both on-policy samples and off-policy samples; (3) Calculating the CARSM gradient estimator; (4) soft updating the target networks for both the policy and critic. The (1) and (3) steps are different from other existing algorithms and we show their pseudo codes in Algorithms \ref{alg1} and \ref{alg3}, respectively.

\begin{algorithm}{
\SetKwData{Left}{left}\SetKwData{This}{this}\SetKwData{Up}{up}
\SetKwFunction{Union}{Union}\SetKwFunction{FindCompress}{FindCompress}
\SetKwInOut{Input}{input}\SetKwInOut{Output}{output}

\textbf{Input:} Policy network $\pi(\av\given\sv,\thetav)$, initial state $\sv_0$, sampled step $T$, replay buffer $\mathcal{R}$\\
\textbf{Output:} Intermediate variable matrix $\varpiv_{1:T}$, logit variables $\phiv_{1:T}$, rewards vector $r_{1:T}$, state vectors $\sv_{1:T}$, action vectors $\av_{1:T}$, replay buffer $\mathcal{R}$
\BlankLine

\For {$t =1 \cdots T$}{
Generate Dirichlet random variable $\varpiv_{tk} \sim \text{Dir}(\mathbf{1}_C)$ for each dimension $k$;

Calculate logits $\phiv_{t} = \mathcal{T}_{\thetav}(\sv_t)$ which is a $K\times C$ length vector

Select action $a_{tk} = \text{argmin}_{i\in\{1,\cdots,C\}}(\ln \varpi_{tki}{-\phi_{tki}})$ for each dimension $k$;

Obtain next state values $\sv_{t+1}$ and reward $r_t$ based actions $\av_t=(a_{t1},\ldots,a_{tK})'$ and current state $\sv_t$.

Store the transition $\{\sv_t,\av_t,r_t,\sv_{t+1}\}$ to replay buffer $\mathcal{R}$

Assign $\sv_t\leftarrow \sv_{t+1}$.
}
}
\caption{Collecting samples from environment}
\label{alg1}
\end{algorithm}

{\begin{algorithm}[h]{\small
\SetKwData{Left}{left}\SetKwData{This}{this}\SetKwData{Up}{up}
\SetKwFunction{Union}{Union}\SetKwFunction{FindCompress}{FindCompress}
\SetKwInOut{Input}{input}\SetKwInOut{Output}{output}

\textbf{Input:} Critic network $Q_{\omegav}$, policy network $\pi_{\thetav}$, on-policy samples including states $\sv_{1:T}$, actions $\av_{1:T}$, intermediate Dirichlet random variables $\varpiv_{1:T}$, logits vectors $\phiv_{1:T}$, discounted cumulative rewards $y_{1:T}$.\\
\textbf{Output:} an updated policy network
\BlankLine

Initialize $g\in\mathbb{R}^{T\times K\times C}$;

\For {t = 1$\cdots$ T (in parallel)}{

\For {k = 1$\cdots$ K (in parallel)}{
Let $A_{tk} = \{(c,j)\}_{{c}=1:{C},~j<c}$ , and initialize $P^{tk}\in\mathbb{R}^{C\times C}$ with all element equals to $a_{tk}$ (true action).

 \For{ $(c,j) \in A_{tk} 
 $ (in parallel)}{
Let $ a_{tk}^{_{c \leftrightharpoons j}}
=\argmin\nolimits_{i\in\{1,\ldots,C_k\}} (\ln \varpi_{tki}^{_{{c} \leftrightharpoons j}} {-\phi_{tki}})
$

\If{$a_{tk}^{_{c \leftrightharpoons j}}$ not equals to $a_{tk}$}{
Assign $a_{tk}^{_{c \leftrightharpoons j}}$ to $P^{tk}(c,j)$
}
}
}
Let $S_{t}=\mbox{unique}(P^{t1}\otimes P^{t2}\cdots\otimes P^{tK})\backslash \{a_{t1}\otimes a_{t2}\cdots\otimes a_{tK}\}$, which means $S_{t}$ is the set of all unique values across $K$ dimensions except for true action $\av_t = \{a_{t1}\otimes a_{t2}\cdots\otimes a_{tK}\}$;
denote pseudo action of swapping between coordinate $c$ and $j$ as $S_t(c,j) = (P^{t1}(c,j) \otimes P^{t1}(c,j)\cdots\otimes P^{tK}(c,j))$, and define $\mathcal{I}_t$ as unique pairs contained in~$S_t$.

\bigskip

Initialize matrix $F^t\in\mathbb{R}^{C\times C}$ with all elements equal to $y_t$;

\For {$(\tilde{c},\tilde{j})\in \mathcal{I}_{t}$ (in parallel)}{
$F^{t}(\tilde{c},\tilde{j}) = Q_{\omegav}(\sv_t, S_t(\tilde{c},\tilde{j})) $
}
Plug in number for matrix $g_{tkc} = \sum_{j=1}^C (F^{t}_c - \bar{F}^{t}_c)(\frac{1}{C}-\varpi_{tkj})$, where $F^{tk}_c$ denotes the $c$th row of matrix $F^{t}$ and $\bar{F}^t_c$ is the mean of that row; 

\For {$k = 1\cdots K$}{
\If{every element in $P^{tk}$ is $a_{tk}$}
{
$g_{tkc} = 0$
}
}
}
Update the parameter for $\thetav$ for policy network by maximize the function 
$$J= \frac{1}{TKC}\sum_{t=1}^T\sum_{k=1}^K\sum_{c=1}^C g_{tkc}\phi_{tkc}$$ 
where $\phi_{tkc}$ are logits and $g_{tkc}$ are placeholders that stop any gradients, and use auto-differentiation on $\phi_{tkc}$ to obtain gradient with respect to $\thetav$. 
}%
\caption{CARSM policy gradient for a $K$-dimensional $C$-way categorical action space.}
\label{alg3}%
\end{algorithm}\vspace{-2mm}}

\end{document}